\documentclass[journal]{IEEEtran}
\usepackage{blindtext}
\usepackage{graphicx}
\usepackage{array}
\usepackage[numbers,sort&compress]{natbib}
\usepackage{amsmath}
\usepackage{color}
\usepackage{algorithm2e}
\usepackage{subcaption}
\let\labelindent\relax
\usepackage{enumitem}
\usepackage{url}
\usepackage{multirow}
\usepackage{moresize}
\usepackage[table]{xcolor}
\usepackage{color,soul}
\usepackage[]{hyperref}
\soulregister\cite7
\soulregister\ref7
\soulregister\pageref7
\usepackage{comment}
\usepackage{dblfloatfix}    
\usepackage[flushleft]{threeparttable} 
\usepackage{xcolor}
\usepackage[geometry]{ifsym}

\begin{document}

\twocolumn[{%
\vspace{40mm}
{ \large
\begin{itemize}[leftmargin=2.5cm, align=parleft, labelsep=2.0cm, itemsep=4ex,]

\item[\textbf{Citation}]{D. Temel, T. Alshawi, M-H. Chen, and G. AlRegib, "Challenging Environments for Traffic Sign Detection: Reliability Assessment under Inclement Conditions," in arXiv:1902.06857, 2019.}

\item[\textbf{Dataset}]{\url{https://github.com/olivesgatech/CURE-TSD}}

\item[\textbf{Bib}]  {@INPROCEEDINGS\{Temel2019\_CURE-TSD,\\ 
author=\{D. Temel and T. Alshawi and M-H. Chen and G. AlRegib\},\\ 
booktitle=\{arXiv:1902.06857\},\\
title=\{Challenging Environments for Traffic Sign Detection:
Reliability Assessment under Inclement Conditions\},\\ 
year=\{2019\}, \} 
}

\item[\textbf{Contact}]{\href{mailto:alregib@gatech.edu}{alregib@gatech.edu}~~~~~~~\url{https://ghassanalregib.info/}\\ \href{mailto:dcantemel@gmail.com}{dcantemel@gmail.com}~~~~~~~\url{http://cantemel.com/}}
\end{itemize}
\thispagestyle{empty}
\newpage
\clearpage
\setcounter{page}{1}
}
}]

\title{Challenging Environments for Traffic Sign Detection: Reliability Assessment under Inclement Conditions}

\author{Dogancan~Temel,~Tariq~Alshawi*,~Min-Hung~Chen*,~and~Ghassan~AlRegib

{\includegraphics[width=0.19\linewidth]{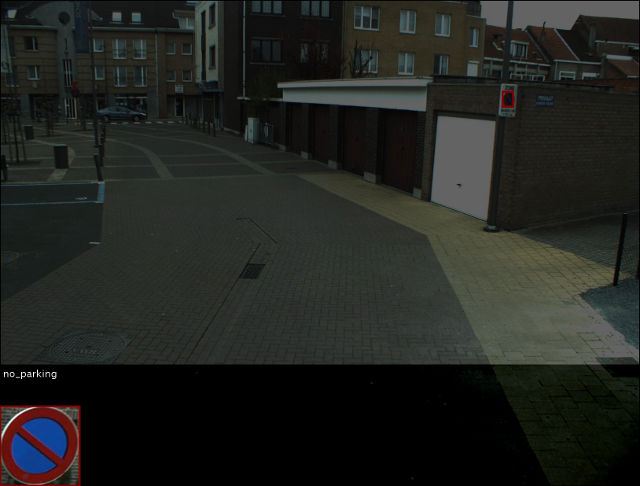}}
{\includegraphics[width=0.19\linewidth]{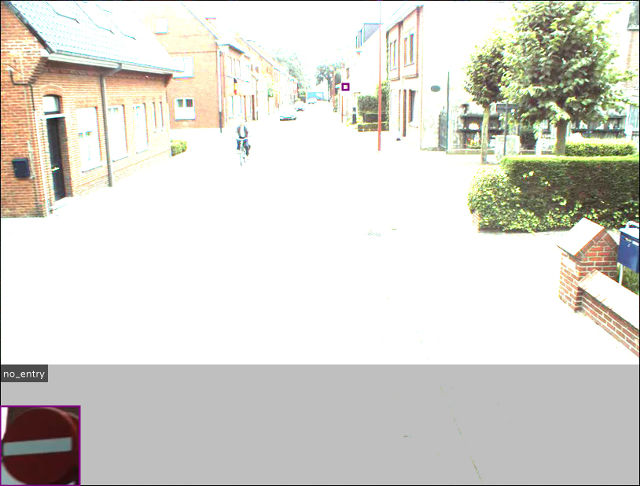}}
{\includegraphics[width=0.19\linewidth]{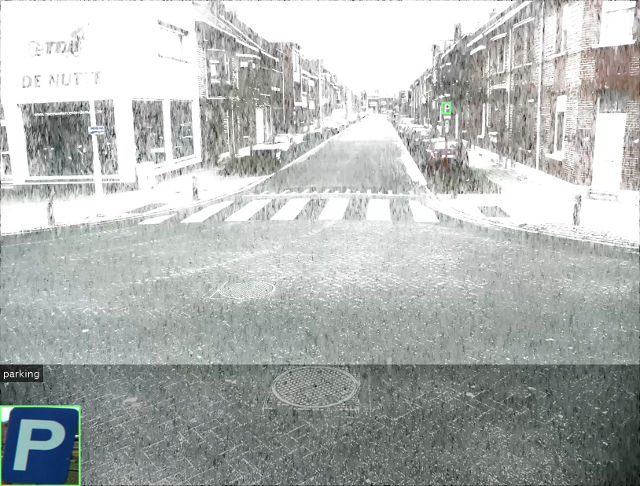}} 
{\includegraphics[width=0.19\linewidth]{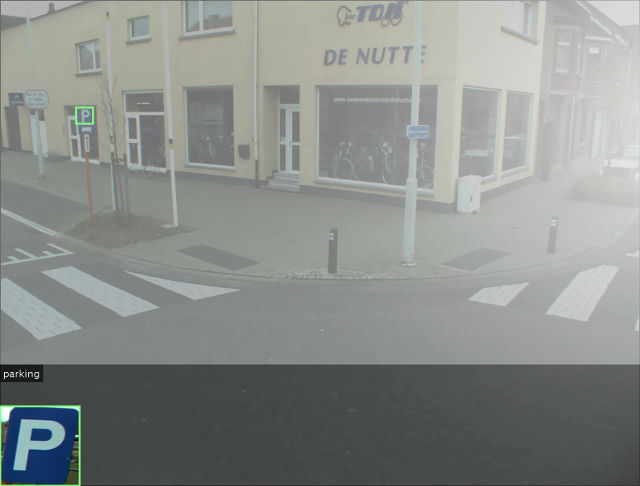}}
{\includegraphics[width=0.19\linewidth]{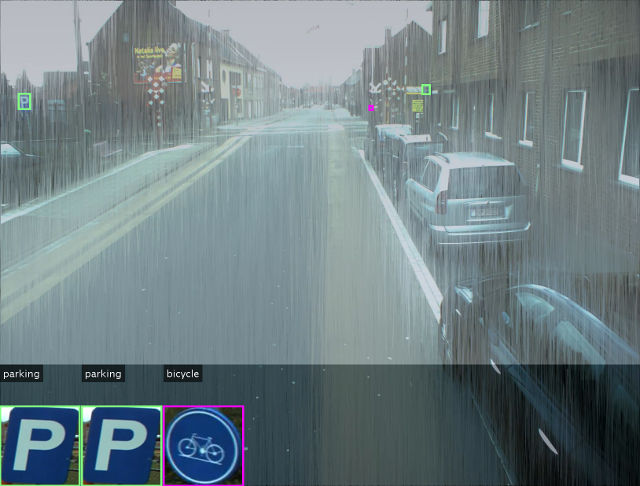}}
\captionof{figure}{Challenging scene examples from the \texttt{CURE-TSD} dataset: darkening, exposure, snow, haze, and rain, respectively.}
\label{fig:intro_image}
\vspace{-10mm}

\thanks{* Equal contribution. Manuscript received July, 2019.  Dogancan Temel, Min-Hung Chen, and Ghassan AlRegib are with the School of Electrical and Computer Engineering, Georgia Institute of Technology, Atlanta,
GA, 30332-0250 USA. Tariq Alshawi is with the Electrical Engineering Department, College of Engineering, King Saud Univeristy, Riyadh, Saudi Arabia. Corresponding author email: cantemel@gatech.edu.}
}

\markboth{}%
{Shell \MakeLowercase{\textit{et al.}}: Bare Demo of IEEEtran.cls for Journals}

\maketitle

\begin{abstract}
State-of-the-art algorithms successfully localize and recognize traffic signs over existing datasets, which are limited in terms of challenging condition type and severity. Therefore, it is not possible to estimate the performance of traffic sign detection algorithms under overlooked challenging conditions. Another shortcoming of existing datasets is the limited utilization of temporal information and the unavailability of consecutive frames and annotations. To overcome these shortcomings, we generated the CURE-TSD video dataset and hosted the first IEEE Video and Image Processing Cup within the IEEE Signal Processing Society. In this paper, we provide a detailed description of the CURE-TSD dataset, analyze the characteristics of the top performing algorithms, and provide a performance benchmark. Moreover, we investigate the robustness of the benchmarked algorithms with respect to sign size, challenge type and severity. Benchmarked algorithms are based on state-of-the-art and custom convolutional neural networks that achieved a precision of $0.55$ and a recall of $0.32$, $F_{0.5}$ score of $0.48$ and $F_{2}$ score of $0.35$. Experimental results show that benchmarked algorithms are highly sensitive to tested challenging conditions, which result in an average performance drop of $0.17$ in terms of precision and a performance drop of $0.28$ in recall under severe conditions. The dataset is publicly available at  \href{https://github.com/olivesgatech/CURE-TSD}{https://github.com/olivesgatech/CURE-TSD}.

\end{abstract}


\IEEEpeerreviewmaketitle

\vspace{-4.0mm}
\section{Introduction}
The reliability of autonomous driving systems depends on the reliability of the core technologies that process and analyze sensed information.  Therefore, algorithmic solutions behind the core technologies have to achieve a minimum performance under challenging conditions. Although extensive test driving have been a standard in the industry, there is an increase in demand for simulated approaches to evaluate the performance of advanced driver-assistence systems and autonomous functionalities. An essential component of simulations is the development of traffic datasets  \cite{Grigorescu2003,Timofte2009,Timofte2014,Belaroussi2010,Larsson2011,Stallkamp2011,Stallkamp2012,Houben2013,Mogelmose2012,Zhu2016,Yang2016,Zhang2017}. The focus of this paper is traffic sign detection that can only be sensed and processed through a vision system.

M{\o}gelmose \textit{et al.} \cite{Mogelmose2012} conducted a traffic sign detection survey, which included German traffic sign recognition benchmark \cite{Stallkamp2011,Stallkamp2012}, Swedish traffic signs dataset \cite{Larsson2011}, RUG traffic sign image dataset \cite{Grigorescu2003}, Stereopolis dataset \cite{Belaroussi2010}, and Belgium traffic sign dataset \cite{Timofte2009}. Moreover in \cite{Mogelmose2012}, the authors also introduced a traffic sign dataset denoted as LISA. In addition to aforementioned datasets, German traffic sign detection benchmark  \cite{Houben2013} and Belgium traffic sign classification \cite{Timofte2014} datasets were also used in the literature \cite{Mathias2013}.  LISA dataset \cite{Mogelmose2012} images were captured in USA whereas other datasets \cite{Grigorescu2003,Timofte2009,Timofte2014,Belaroussi2010,Larsson2011,Stallkamp2011,Stallkamp2012,Houben2013,Zhu2016,Yang2016,Zhang2017} originate from Europe. Even though majority of these datasets were focused on Europe, there are recent datasets that are based on Chinese traffic signs \cite{Zhu2016,Yang2016,Zhang2017}. Zhu \textit{et al.} \cite{Zhu2016} introduced the Tsinghua Tencent 100K dataset, Yang \textit{et al.} \cite{Yang2016} introduced the Chinese Traffic Sign Dataset (CTSD), and Zhang \textit{et al.} \cite{Zhang2017} extended the CTSD dataset with more images from Chinese highways to obtain the Changsha University of Science and Technology Chinese Traffic Sign Detection Benchmark (CCTSDB). 

The majority of incumbent traffic datasets originate from video sequences. However, temporal relationships are overlooked because of discrete labels. Furthermore, these datasets are limited in terms of challenging environmental conditions and lack of performance analysis with respect to these conditions. Recent studies \cite{Lu2017, Das2017} showed that adversarial examples can significantly degrade the traffic sign recognition and detection performance. Even though these recent studies shed a light on the vulnerability of existing methods, analyzed adversarial examples do not necessarily correspond to challenging conditions in the real world. We previously investigated the robustness of algorithms and APIs  under simulated real-world challenging conditions \cite{Temel2017_NIPSW, Temel2018_CUREOR,Temel2019_CUREOR} but these studies were only limited to recognition tasks.

Previously, we organized a competition to assess the reliability of traffic sign detection algorithms and described the VIP Cup experience in a non-technical article, which included an overview of the competition setup and statistics, finalist teams and their opinions \cite{Temel2018_SPM}. In \cite{Temel2019_ITS}, we analyzed the spectral characteristics of the challenging conditions and investigated the relationship between these spectral characteristics and performance variations. In \cite{Temel2017_NIPSW}, we briefly investigated the robustness of recognition algorithms over cropped traffic signs and analyzed two other cropped sign datasets. Cropped traffic sign images were also used in \cite{Prabhushankar2018, Kwon2019} for out-of-distribution classification and robust sign recognition. In this study, we focus on the detection task, which includes localization and recognition, and we provide a comprehensive benchmark of the top performing algorithms in the VIP Cup along with details about the competition dataset. The contributions of this paper are five folds.

\begin{itemize}[label=\textcolor{orange}{\FilledSmallSquare},leftmargin=*]
 \setlength\itemsep{0.5 em}
    \item  We provide a performance benchmark of the top performing algorithms along with algorithmic details for the first time since the competition.  
    \item  We analyze the robustness of benchmark algorithms with respect to challenge type and severity.
    \item We analyze the significance of sign size with respect to traffic sign detection performance. 
    \item  We include a detailed analysis of ten incumbent traffic sign datasets and a comparison with the introduced \texttt{CURE-TSD} dataset, which is the most comprehensive publicly-available traffic sign detection dataset with simulated challenging conditions, to the best of our knowledge. 
     \item We describe the details of ground truth generation process, which is based on a publicly available annotation tool and a simulator engine. Moreover, we briefly describe the challenging condition synthesis framework.
    \end{itemize}

\begin{center}
\begin{table*}[htbp!]
\ssmall
\centering
\caption{Main characteristics of publicly available datasets and CURE-TSD dataset}
\label{tab_datasets}

\begin{threeparttable}

{\renewcommand{\arraystretch}{1.2}

\begin{tabular}{>{\centering\arraybackslash}p{1.25cm}>{\centering\arraybackslash}p{1.05cm}>{\centering\arraybackslash}p{1.02cm}>{\centering\arraybackslash}p{0.93cm}>{\centering\arraybackslash}p{1.10cm}>{\centering\arraybackslash}p{1.2cm}>{\centering\arraybackslash}p{1.25cm}>{\centering\arraybackslash}p{1.0cm}>{\centering\arraybackslash}p{1.10cm}>{\centering\arraybackslash}p{1.15cm}>{\centering\arraybackslash}p{1.05cm}>{\centering\arraybackslash}p{1.15cm}}
\hline
 & \textbf{\href{http://btsd.ethz.ch/shareddata/}{BelgiumTS}} & \textbf{\href{http://btsd.ethz.ch/shareddata/}{BelgiumTSC}} & \textbf{\href{http://www.itowns.fr/roadsign.php}{Stereopolis}} & \textbf{\href{http://www.cvl.isy.liu.se/en/research/datasets/traffic-signs-dataset/}{STS}} &\textbf{\href{http://benchmark.ini.rub.de/}{GTSRB}}  & \textbf{\href{http://cvrr.ucsd.edu/LISA/lisa-traffic-sign-dataset.html}{LISA}}  & \textbf{\href{http://benchmark.ini.rub.de/}{GTSDB}}  & \textbf{\href{http://cg.cs.tsinghua.edu.cn/traffic-sign/}{TT-100K}}  & \textbf{\href{http://luo.hengliang.me/data.htm}{CTSD}} & \textbf{\href{https://github.com/csust7zhangjm/CCTSDB}{CCTSDB}}  & \textbf{CURE-TSD} \\ \hline

\textbf{Release year}  & 2009 & 2009 & 2010 & 2011 & 2011 & 2012 & 2013 &2016 & 2016 &2017 Sept. & 2017 March \\ \hline


\textbf{\begin{tabular}[p]{@{}c@{}}Number of  \\ video \\sequences\end{tabular}}  & 4 & N/A & N/A & N/A & N/A & \begin{tabular}[c]{@{}c@{}}17\\ (only tracks \\are available) \end{tabular} & N/A &N/A &N/A &N/A & 5,733 \\ \hline

\textbf{\begin{tabular}[p]{@{}c@{}}Number of \\frames\\ per video\end{tabular}}  & \begin{tabular}[c]{@{}c@{}}2,001 to\\ 6,201\end{tabular} & N/A & N/A & N/A & N/A & \begin{tabular}[c]{@{}c@{}}up to 30\\ per track\end{tabular}  & N/A &N/A &N/A &N/A  & 300 \\ \hline

\textbf{\begin{tabular}[c]{@{}c@{}}Number of \\ frames/images\end{tabular}}  & 15,204 & 7,125 & 847 & 20,000 & \begin{tabular}[c]{@{}c@{}}144,769\end{tabular} & 6,610 & 900 &100,000 &1,100 &10,000 & 1,719,900 \\ \hline

\textbf{\begin{tabular}[c]{@{}c@{}}Number of \\annotated\\ frames/images\end{tabular}}  & 9,006 & 7,125 & 273 & 3,488 & 51,840 & \begin{tabular}[c]{@{}c@{}}All\\ frames\end{tabular} & \begin{tabular}[c]{@{}c@{}}All\\ frames\end{tabular} &\begin{tabular}[c]{@{}c@{}}All\\ frames\end{tabular} & \begin{tabular}[c]{@{}c@{}}All\\ frames\end{tabular} &\begin{tabular}[c]{@{}c@{}}All\\ frames\end{tabular} & \begin{tabular}[c]{@{}c@{}}All\\ frames\end{tabular} \\ \hline

\textbf{\begin{tabular}[c]{@{}c@{}}Number of \\ annotated signs\end{tabular}}  & 13,444 & 7,125 & 273 & 3,488 & 51,840 & 7,855 & 1,206 &30,000 & 1,574 &13,361 & 2,206,106 \\ \hline

\textbf{\begin{tabular}[p]{@{}c@{}}Annotation\\ information\end{tabular}}  & \begin{tabular}[c]{@{}c@{}}Sign type, \\ bounding box, \\3D location\end{tabular} & \begin{tabular}[c]{@{}c@{}}Sign type, \\ bounding box \end{tabular} &\begin{tabular}[c]{@{}c@{}}Sign type,\\ bounding \\ box\end{tabular} & \begin{tabular}[c]{@{}c@{}}Sign type,  \\ boudning box,\\ visibility and \\ road status\end{tabular} & Sign type & \begin{tabular}[c]{@{}c@{}}Sign type,  \\bounding box,\\occlusion and\\ road status\end{tabular} & \begin{tabular}[c]{@{}c@{}}Sign type,\\bounding box\end{tabular} &\begin{tabular}[c]{@{}c@{}}Sign type,  \\bounding box, \\ pixel map \end{tabular} &\begin{tabular}[c]{@{}c@{}}Sign type,  \\bounding box \end{tabular} &\begin{tabular}[c]{@{}c@{}}Sign category,  \\bounding box \end{tabular}  & \begin{tabular}[c]{@{}c@{}}Sign type,  \\bounding box, \\ challenge type \\ and level\end{tabular} \\ \hline


\textbf{Resolution}  & 1,628x1,236 &  \begin{tabular}[c]{@{}c@{}}22x21 to \\ 674x527\end{tabular} & 1,920x1,080 & 1,280x960 & \begin{tabular}[c]{@{}c@{}}15x15 to \\ 250x250\end{tabular} & \begin{tabular}[c]{@{}c@{}}640x480 to\\ 1,024x522\end{tabular} & 1,360x800 &2,048x2,048 &\begin{tabular}[c]{@{}c@{}}1,024x768 \&\\ 1,270x800\end{tabular} &\begin{tabular}[c]{@{}c@{}}1,024x768 \&\\ 1,270x800 \&\\ 1,000x350 \end{tabular} & 1,628x1,236 \\ \hline

\textbf{\begin{tabular}[c]{@{}c@{}}Number of \\ sign types\end{tabular}}  & 62 & 62 & 10 & 7 & 43 & 47 & 43 &45 &48 &3 classes  & 14 \\ \hline

\textbf{\begin{tabular}[c]{@{}c@{}}Annotated \\ sign size\end{tabular}}  & \begin{tabular}[c]{@{}c@{}}9x10 to\\ 206x277\end{tabular} & \begin{tabular}[c]{@{}c@{}}11x10 to\\ 562x438\end{tabular} &\begin{tabular}[c]{@{}c@{}}25x25 to\\ 204x159 \end{tabular}  & \begin{tabular}[c]{@{}c@{}}3x5 to\\ 263x248\end{tabular} & \begin{tabular}[c]{@{}c@{}}15x15 to\\ 250x250\end{tabular} & \begin{tabular}[c]{@{}c@{}}6x6 to \\ 167x168\end{tabular} &  \begin{tabular}[c]{@{}c@{}}16-128 \\ (longer \\ edge) \end{tabular} & \begin{tabular}[c]{@{}c@{}}2x7 to \\ 397x394\end{tabular} & \begin{tabular}[c]{@{}c@{}}26x26 to \\ 573x557\\ \end{tabular} & \begin{tabular}[c]{@{}c@{}}10x11 to \\ 380x378\\ \end{tabular} & 
\begin{tabular}[c]{@{}c@{}}3x7 to \\ 206x277\end{tabular} \\  \hline

\textbf{\begin{tabular}[c]{@{}c@{}}Origin of \\ the videos\end{tabular}} & \begin{tabular}[c]{@{}c@{}}Captured \\in\\ Belgium\end{tabular} & \begin{tabular}[c]{@{}c@{}}Captured \\in\\ Belgium\end{tabular} & \begin{tabular}[c]{@{}c@{}}Captured \\in\\ France\end{tabular} & \begin{tabular}[c]{@{}c@{}}Captured \\in\\ Sweeden\end{tabular} & \begin{tabular}[c]{@{}c@{}}Captured \\in\\ Germany\end{tabular} & \begin{tabular}[c]{@{}c@{}}Captured \\in\\ California\end{tabular} & \begin{tabular}[c]{@{}c@{}}Captured \\in\\ Germany\end{tabular} & \begin{tabular}[c]{@{}c@{}}Captured \\in\\ China\end{tabular}  & \begin{tabular}[c]{@{}c@{}}Captured \\in\\ China\end{tabular} &\begin{tabular}[c]{@{}c@{}}CTSD \\and\\ China\end{tabular} & \begin{tabular}[c]{@{}c@{}}BelgiumTS, \\  Unreal Engine, \\ Adobe After \\ Effects \end{tabular} \\ \hline

\textbf{\begin{tabular}[c]{@{}c@{}}Acquisition \\ device\end{tabular}}  & \begin{tabular}[c]{@{}c@{}}Color\\ cameras\end{tabular} & \begin{tabular}[c]{@{}c@{}}Color \\ cameras\end{tabular} & \begin{tabular}[c]{@{}c@{}}Color\\ cameras\end{tabular} & \begin{tabular}[c]{@{}c@{}}Point-Grey\\ Chameleon\\ color camera\end{tabular} & \begin{tabular}[c]{@{}c@{}}Prosilica GC\\ 1380CH\\ color camera\end{tabular} & \begin{tabular}[c]{@{}c@{}}Color and \\ grayscale\\ cameras\end{tabular} & \begin{tabular}[c]{@{}c@{}}Prosilica GC\\ 1380CH\\ color camera\end{tabular} &  \begin{tabular}[c]{@{}c@{}}Color \\ cameras\end{tabular} &\begin{tabular}[c]{@{}c@{}}Color\\ cameras\end{tabular} &\begin{tabular}[c]{@{}c@{}}Color\\ cameras\end{tabular} & \begin{tabular}[c]{@{}c@{}}BelgiumTS: \\ Proscilica GC\\ 1380CH\end{tabular} \\ \hline


\textbf{\begin{tabular}[c]{@{}c@{}}Training-test \\ data splitting \end{tabular}}  & 65\%-35\% & 65\%-35\% & 100\% test & \begin{tabular}[c]{@{}c@{}}2 test sets, \\ no splitting\end{tabular} & \begin{tabular}[c]{@{}c@{}}Random \\ splitting, \\ validation:25\%,\\training:50, \\  test:25\%\end{tabular} & \begin{tabular}[c]{@{}c@{}}Random \\ splitting,\\ 80\%-20\%\end{tabular} & \begin{tabular}[c]{@{}c@{}}Random \\splitting,\\ 66.6\%-33.4\% \end{tabular} &\begin{tabular}[c]{@{}c@{}}Random \\splitting,\\ 66.6\%-33.4\% \end{tabular} & \begin{tabular}[c]{@{}c@{}}Random \\splitting,\\ 63.6\%-36.4\% \end{tabular} &\begin{tabular}[c]{@{}c@{}}Random \\splitting \end{tabular} & \begin{tabular}[c]{@{}c@{}}Statistical\\ splitting,\\ training:70\%,\\ test:30\%\end{tabular} \\ \hline

\textbf{\begin{tabular}[c]{@{}c@{}}Challenging \\ conditions \\ (*=annotated) \end{tabular}}   
&\begin{tabular}[c]{@{}c@{}}illumination,\\occlusion,\\ overcast \end{tabular}  
&\begin{tabular}[c]{@{}c@{}} illumination,\\occlusion,\\deformation,\\ perspective, \\size  \end{tabular}  
&  \begin{tabular}[c]{@{}c@{}}illumination,\\occlusion,\\overcast\end{tabular} 
& \begin{tabular}[c]{@{}c@{}}illumination,\\occlusion*,\\ shadow, blur*,\\overcast, rain
\end{tabular} 
&\begin{tabular}[c]{@{}c@{}} illumination,\\occlusion,\\deformation,\\size, blur \\perspective   \end{tabular} 
&\begin{tabular}[c]{@{}c@{}}\\ illumination,\\occlusion*,\\ shadow, blur,\\ reflection,\\codec error,\\dirty lens,\\ overcast  \end{tabular}
&\begin{tabular}[c]{@{}c@{}}illumination,\\occlusion,\\blur, shadow,\\haze, rain,\\overcast  \end{tabular}
& \begin{tabular}[c]{@{}c@{}}illumination, \\occlusion, \\overcast,\\haze, shadow \end{tabular}   
&\begin{tabular}[c]{@{}c@{}}illumination,\\occlusion,\\shadow, rain,\\overcast,\\dirty lens,\\reflection,\\haze,blur  \end{tabular}
&\begin{tabular}[c]{@{}c@{}}illumination,\\occlusion,\\shadow, rain,\\overcast, \\dirty lens,\\reflection,\\haze, blur \end{tabular} &\begin{tabular}[c]{@{}c@{}}rain*, snow*, \\ shadow*, haze*,\\ illumination* , \\decolorization*,\\blur*, noise*,\\codec error*,\\dirty lens*,\\occlusion, \\overcast \end{tabular} \\ \hline
\end{tabular}
}
\end{threeparttable}
\end{table*}
\end{center}
\vspace{-6.0mm}
\section{Related Work}
Robust and reliable traffic sign detection is necessary to bring autonomous vehicles onto our roads. Therefore, researchers have been heavily focused on designing traffic sign detection and recognition algorithms. Initially, fully handcrafted methods dominated the traffic sign literature \cite{Gavrila1999,Grigorescu2003,Belaroussi2010,Larsson2011}. The main mechanisms in these methods can be divided into two main blocks. First block is based on extracting descriptive information, which can be in the form of a feature, a descriptor, or any other representation. Second block is based on assessing the similarity/dissimilarity between extracted representations and predefined representations such as templates to determine the class of a traffic sign. Even though these handcrafted methods can lead to high detection or recognition performance in small scale datasets such as RUG \cite{Grigorescu2003}, state-of-the-art methods were dominated by a wide range of machine learning methods in more recent datasets  \cite{Bahlmann2005,Bascon2007,Broggi2007,Timofte2009,Stallkamp2011,Greenhalgh2012,Mogelmose2012,Berkaya2016,Ellahyani2016, Yuan2017}. The machine learning techniques that lead to state-of-the-art performances in these studies include AdaBoost, neural networks, support vector machines, linear discriminant analysis, subspace analysis, ensemble classifiers, slow feature analysis, kd-trees, and random forests. Aforementioned studies mainly utilize machine learning techniques to learn the correspondence between handcrafted features/descriptors and traffic sign classes. However, Convolutional neural networks (CNNs) enabled learning visual representations in conjunction with their correspondences to sign classes directly from data, which led to state-of-the-art recognition and detection performances in recent studies \cite{Sermanet2011,Jin2014,Zhu2016Neuro,Aghdam2016,Zhu2016,Li2016,Zeng2017,Hu2017,Zhang2017,Serna2018,Wong2018}. Even though CNNs are commanly used to eliminate designing features, they can also be combined with handcrafted features as shown in \cite{Yang2016,Luo2017}.

There are numerous publicly-available datasets in the literature to validate traffic sign detection and recognition algorithms. The characteristics of these datasets are summarized in Table \ref{tab_datasets}, which includes release year, number of video sequences, number of frames per video, number of total frames/images, number of total annotated frames/images, number of annotated signs, annotation information, resolution, number of sign types, annotated sign size, origin of the videos, acquisition device, training/test data splitting information, and challenging conditions. We have reported the characteristics of publicly available datasets based on either the information within the reference papers or our own analysis of dataset files. In the dataset table, there is a not-applicable expression ($N/A$) when a category does not apply to a dataset. Additional information about datasets are given in Section \ref{subsec_existing_datasets}, main characteristics of these datasets are summarized in Section \ref{subsec_datasets_main}, and shortcomings are described in Section \ref{subsec_datasets_short}.

\subsection{Existing Datasets}
\label{subsec_existing_datasets}

\noindent \textbf{RUG \cite{Grigorescu2003}:} Grigorescu \textit{et al.} \cite{Grigorescu2003} utilized distance sets for shape recognition, which also includes traffic signs. As test set, they introduced the \texttt{\href{http://www.cs.rug.nl/~imaging/databases/traffic_sign_database/traffic_sign_database.html}{RUG}} dataset that has $48$ images of three traffic sign types captured in Netherlands with color cameras. Images have a resolution of $360 \times 270$ and stored as PNG files.

\vspace{2.0 mm}

\noindent \textbf{BelgiumTS \cite{Timofte2009} and BelgiumTSC \cite{Timofte2014}:} Timofte \textit{et al.} \cite{Timofte2009} generated the BelgiumTS dataset, which was created with a van that had 8 roof-mounted cameras capturing images every meter. BelgiumTS includes a training set, a 2D testing set, and a 3D testing set. A subset of BelgiumTS images were selected to create Belgium traffic sign classification (BelgiumTSC) dataset in \cite{Timofte2014}.

\vspace{2.0 mm}

\noindent \textbf{Stereopolis \cite{Belaroussi2010}:} Belaroussi \textit{et al.} \cite{Belaroussi2010} introduced the Stereopolis dataset, which contains images of complex urban scenes that were captured with a van every five minutes.

\vspace{2.0 mm}

\noindent \textbf{STS \cite{Larsson2011}:} Larsson and Felsberg \cite{Larsson2011} generated the STS dataset. In the acquisition process, recording was started by a human operator whenever a traffic sign was visible and stopped when it was out of sight. In the annotations, visibility status includes visible, blurred, and occluded, and road status includes current road or side road.
 
\vspace{2.0 mm}

\noindent \textbf{GTSRB \cite{Stallkamp2011,Stallkamp2012} and GTSDB \cite{Houben2013} :} Stallkamp \textit{et al.} \cite{Stallkamp2011,Stallkamp2012} introduced the German traffic sign recognition benchmark (GTSRB) dataset. The sequence of images originating from one traffic sign instance was registered as track, which included 30 images by design. The training set maintained the temporal order of the images whereas the images in the test set were shuffled. The test set used for online competition \cite{Stallkamp2011} was later utilized as the validation set in \cite{Stallkamp2012}. German traffic sign detection benchmark (GTSDB) \cite{Houben2013} is a successor to the GTSRB.  In training/test data splitting, images of same real world traffic sign was assigned to either training or test set. 

\vspace{2.0 mm}

\noindent \textbf{LISA \cite{Mogelmose2012} :} M{\o}gelmose \textit{et al.} \cite{Mogelmose2012} introduced the LISA dataset, which was acquired with various systems including an Audi vehicle with Intelligent Urban Assist technology, a bike, and a Canon S95. After acquisition, each source video was split into smaller videos (tracks) that include up to $30$ equidistant frames and annotated frames are at least $5$ frames apart.

 \vspace{2.0 mm}

\noindent \textbf{Tencent 100K \cite{Zhu2016} :} Zhu \textit{et al.} \cite{Zhu2016} introduced the Tsinghua-Tencent 100K dataset by extracting data from Tencent Street Views, which were captured with $6$ SLR cameras every 10 meters and stitched together. The top and the bottom $25\%$ of panoramic images were cropped and remaining images were divided vertically into $4$ sub-images. Polygon and ellipse shapes were utilized to obtain pixel maps for traffic signs.

 \vspace{2.0 mm}

 \noindent \textbf{CTSD \cite{Yang2016} and CCTSDB \cite{Zhang2017} :} Yang \textit{et al.} \cite{Yang2016} introduced the Chinese Traffic Sign Dataset (CTSD), which includes images captured on highway, urban, and rural roads. Zhang \textit{et al.} \cite{Zhang2017} introduced the Changsha University of Science and Technology Chinese Traffic Sign Detection Benchmark (CCTSDB) by extending the CTSD dataset with $5,200$ images captured in Chinese highways. Moreover, they augmented the original images by adding noise, changing illumination, and scaling to increase the diversity in the training of their models.

\subsection{Main Characteristics of Existing Datasets}
\label{subsec_datasets_main}
The majority of datasets summarized in Table \ref{tab_datasets} include images or groups of images denoted as tracks. There are $17$ tracks in the LISA dataset, which includes ordered frames and the maximum number of frames in these tracks is $30$. In addition to the LISA dataset, BelgiumTS dataset also includes ordered frames that correspond to $4$ video sequences. The number of frames in these sequences vary from $2,001$ to $6,201$. The number of frames/images in existing datasets is in between $48$ and $144,769$ and the number of annotated signs in these datasets vary from none to $51,840$. The annotation information in these datasets can include sign type, bounding box, 3D location, pixel map, visibility, occlusion, and road status. Bounding boxes were selected with the 
\texttt{\href{http://cvrr.ucsd.edu/LISA/lisa-traffic-sign-dataset.html}{Video Annotator and Frame Annotator}} tool  in the LISA dataset  and the \texttt{\href{http://www.lg.com/hk\_en/ultraHD/index}{NISYS}} tool in the GTSRB and GTSDB datasets. The resolution of full-scene images in the analyzed datasets vary from $360\times270$ to $2,048\times2,048$ and the annotated sign size vary from $2\times7$ to $573\times557$. We have analyzed the images along with the metadata in these datasets to obtain the dataset characteristics when they are not available in manuscripts. We have observed that some of the characteristics reported in corresponding papers including number of images, min/max resolution, and annotation size can slightly differ from the materials that are publicly available online.

Challenging conditions in traffic datasets are not explicitly described or analyzed in most of the corresponding publications. Therefore, we have visually inspected the images in these datasets to summarize existing conditions. Challenging conditions in detection datasets include illumination, occlusion, shadow, blur, reflection, codec error, dirty lens, overcast, and haze. In recognition datasets, we observe micro-level challenging conditions including perspective, size and deformation of the signs. The number of traffic sign types included in these datasets is in between $3$ and $62$. The majority of prior datasets were captured in Europe including Netherlands, Belgium, France, Sweeden, and Germany whereas LISA was captured in the USA and more recent datasets including TT-100K, CTSD, and CCTSDB were captured in China. Color cameras were utilized in the acquisition of majority of these datasets. However, grayscale cameras can be preferred to color cameras in mass production of cars to minimize overall cost. In the LISA dataset, some of the images were directly obtained from the sensors of an Audi A8 vehicle, which provided grayscale images. In some of the datasets, there is no split for training and testing whereas in other datasets, images were randomly split while satisfying predefined conditions.


\subsection{Shortcomings of Existing Datasets}
\label{subsec_datasets_short}
The shortcomings of existing traffic sign datasets are five folds. First, the majority of existing datasets are limited in terms of challenging environmental conditions. Even though some of these datasets include numerous challenging conditions, these conditions are usually limited to marginal levels and they do not include severe conditions. Second, even some of the sequences in these datasets do have challenging conditions, corresponding metadata including condition type and level are not provided. Third, it is not possible to perform controlled experiments based on solely acquired traffic data because there is limited control over the conditions. Fourth, the majority of publicly-available datasets do not include videos and the ones that include videos do not provide labels for all the frames. Fifth, the size of existing datasets may not be sufficient to train models that are based on deep architectures. To compensate the shortcomings of existing datasets in terms of limited challenging conditions and corresponding metadata,  lack of control in the environment, and restricted temporal information, we introduced the \texttt{CURE-TSD} dataset.

\section{CURE-TSD Dataset}
\subsection{Reference Environments}
The majority of publicly available datasets do not provide video sequences as reported in Table \ref{tab_datasets}. However, BelgiumTS \cite{Timofte2009} and LISA \cite{Mogelmose2012} datasets include video sequences. We preferred the BelgiumTS dataset because of the unavailability of videos sequences in the LISA dataset when this study was conducted. There were $8$ cameras used in the acquisition of the BelgiumTS dataset and we utilized the sequences indexed with $03$, which corresponds to the perspective of a driver without any car part blocking the view. In the BelgiumTS dataset, there are four sequences, which include $3,001$, $6,201$, $2,001$, and $4,001$ frames. We generated sequences by combining $300$ frames, which resulted in $49$ challenge-free real-world video sequences. Moreover, we generated synthesized video sequences with a professional game development tool
Unreal Engine 4 (\texttt{UE4}).

We utilized \texttt{UE4} to synthesize video sequences because of three main advantages. First, objects in \texttt{UE4} have metadata including position, size, and bounding box, which eliminates the manual labeling process. Second, environmental conditions including weather, time, and lighting can be fully controlled in \texttt{UE4}. We can change specific environmental conditions and levels while fixing other parameters to perform controlled experiments.  Third, \texttt{UE4} is relatively user friendly and is supported by a large developer community. We started data generation process by creating a virtual city that included roads, street lamps, traffic signs, and background environment. We added global lighting sources to make virtual environment more realistic. We generated a car object and linked that object with two functions, one for obtaining the location and the other one for showing the object on the screen. We attached a camera object to the car object, designed a driving path, created a path follower, configured the car speed, and placed the vehicle on the designed path. To match the number of real-world sequences, we generated $49$ simulated sequences, which leads to a total of $98$ reference videos denoted as challenge-free sequences. 

\newlength{\tempheight}
\newlength{\tempwidth}
\newlength{\vblank}

\newcommand{\columnname}[1]
{\makebox[\tempwidth][c]{\textbf{#1}}}

\newcommand{\rowname}[1]
{\begin{minipage}[b]{0.5cm}
    \centering\rotatebox{90}{\makebox[\tempheight][c]{\textbf{#1}}}
\end{minipage}}

\begin{figure}[!htbp]

    \setlength{\tempwidth}{.305\linewidth}
    \setlength{\vblank}{0mm}
    \settoheight{\tempheight}{\includegraphics[width=\tempwidth]{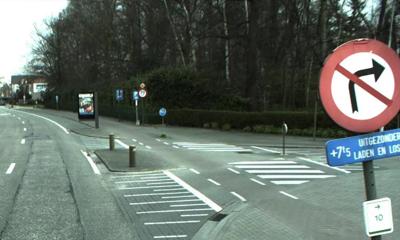}}
    \begin{minipage}[b]{0.5cm}\centering{\makebox[0.5cm][c]{}}\end{minipage}
        \begin{minipage}[c][0.6cm][t]{\tempwidth}\columnname{Level:1}\end{minipage}\hfill
        \begin{minipage}[c][0.6cm][t]{\tempwidth}\columnname{Level:3}\end{minipage}\hfill
        \begin{minipage}[c][0.6cm][t]{\tempwidth}\columnname{Level:5}\end{minipage}
    \rowname{Decolor.}
        {\includegraphics[width=\tempwidth]{./Figs/challenges/01_21_01_01_01_10_0001.jpg}}\vspace{\vblank}\hfill
        {\includegraphics[width=\tempwidth]{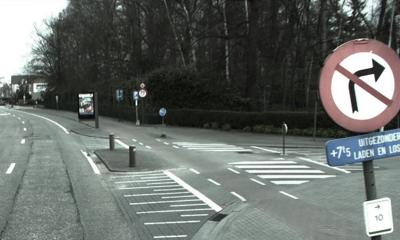}}\hfill
        {\includegraphics[width=\tempwidth]{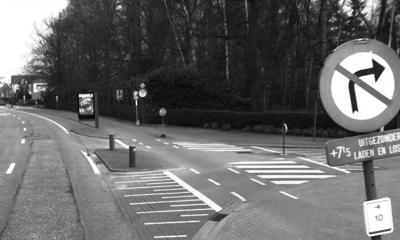}}\\
    \rowname{Lens blur}
        {\includegraphics[width=\tempwidth]{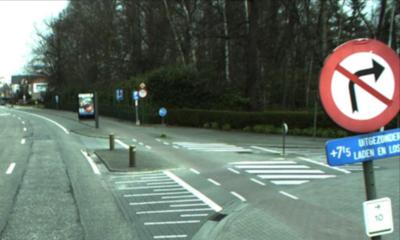}}\vspace{\vblank}\hfill
        {\includegraphics[width=\tempwidth]{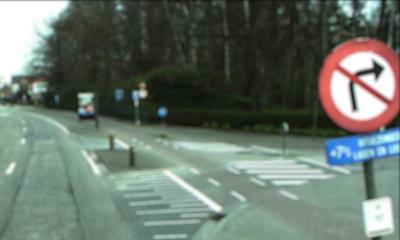}}\hfill
        {\includegraphics[width=\tempwidth]{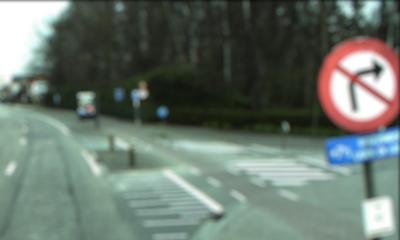}}\\
    \rowname{Codec er.}
        {\includegraphics[width=\tempwidth]{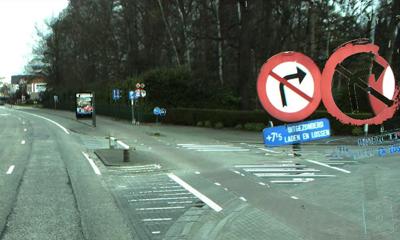}}\vspace{\vblank}\hfill
        {\includegraphics[width=\tempwidth]{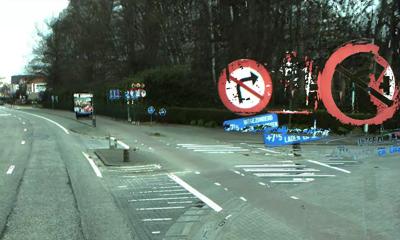}}\hfill
        {\includegraphics[width=\tempwidth]{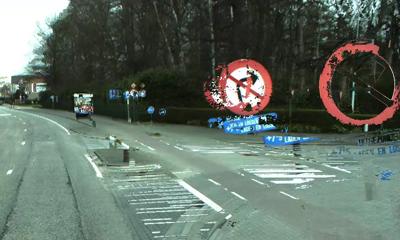}}\\
    \rowname{Darkening}
        {\includegraphics[width=\tempwidth]{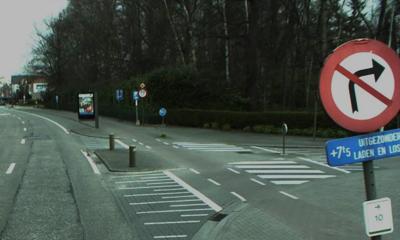}}\vspace{\vblank}\hfill
        {\includegraphics[width=\tempwidth]{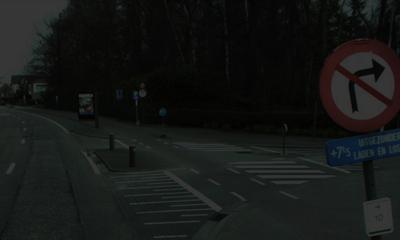}}\hfill
        {\includegraphics[width=\tempwidth]{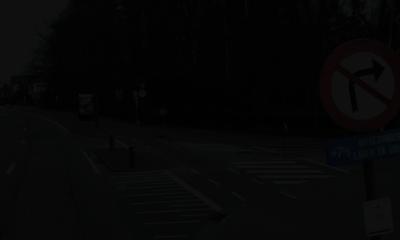}}\\
    \rowname{Dirty lens}
        {\includegraphics[width=\tempwidth]{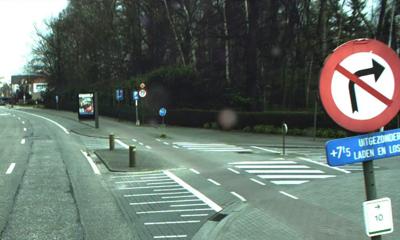}}\vspace{\vblank}\hfill
        {\includegraphics[width=\tempwidth]{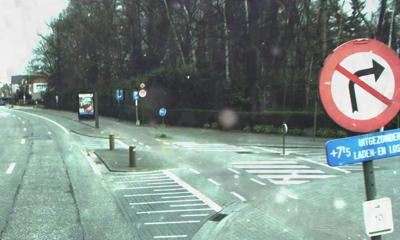}}\hfill
        {\includegraphics[width=\tempwidth]{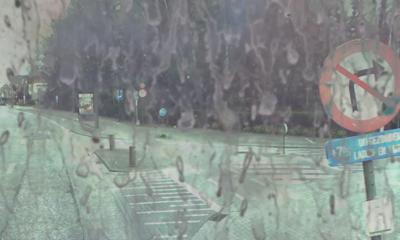}}\\
    \rowname{Exposure}
        {\includegraphics[width=\tempwidth]{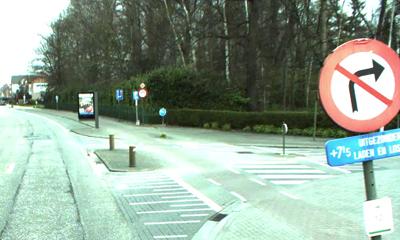}}\vspace{\vblank}\hfill
        {\includegraphics[width=\tempwidth]{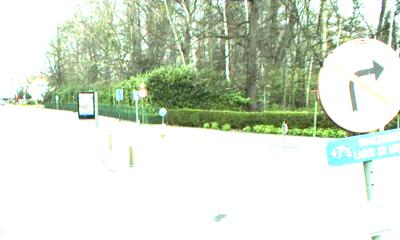}}\hfill
        {\includegraphics[width=\tempwidth]{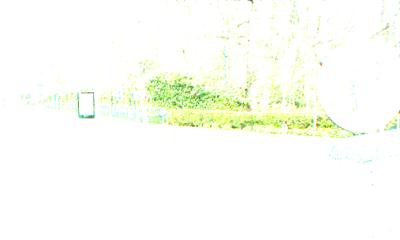}}\\
    \rowname{Gaus. blur}
        {\includegraphics[width=\tempwidth]{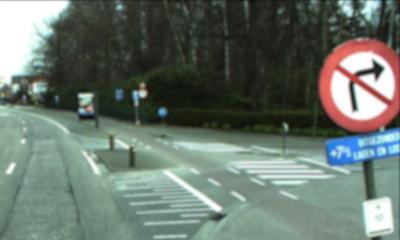}}\vspace{\vblank}\hfill
        {\includegraphics[width=\tempwidth]{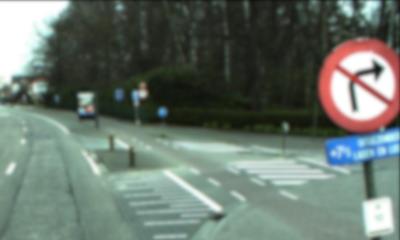}}\hfill
        {\includegraphics[width=\tempwidth]{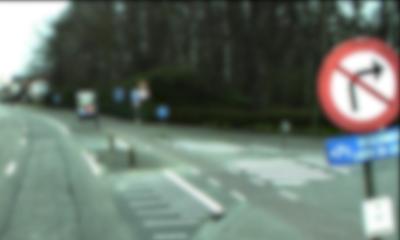}}\\
    \rowname{Noise}
        {\includegraphics[width=\tempwidth]{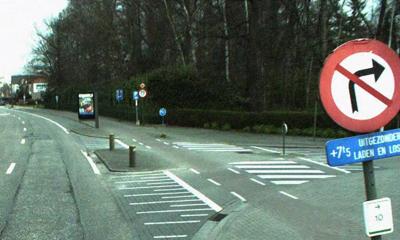}}\vspace{\vblank}\hfill
        {\includegraphics[width=\tempwidth]{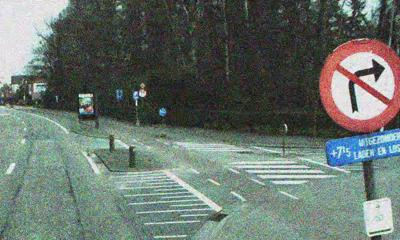}}\hfill
        {\includegraphics[width=\tempwidth]{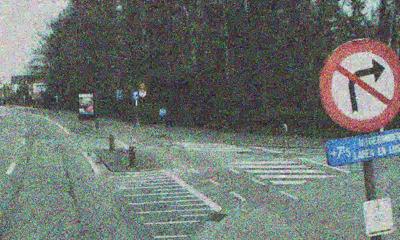}}\\
    \rowname{Rain}
        {\includegraphics[width=\tempwidth]{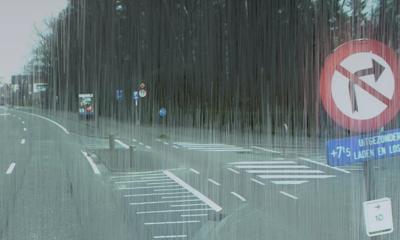}}\vspace{\vblank}\hfill
        {\includegraphics[width=\tempwidth]{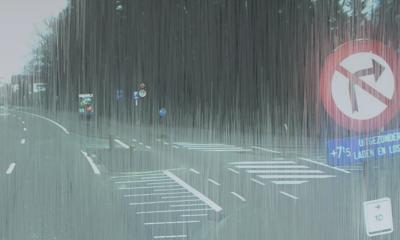}}\hfill
        {\includegraphics[width=\tempwidth]{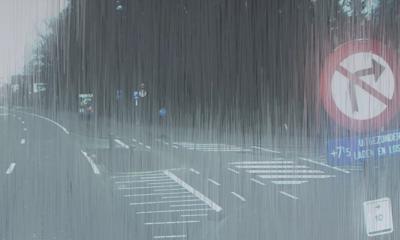}}\\
    \rowname{Shadow}
        {\includegraphics[width=\tempwidth]{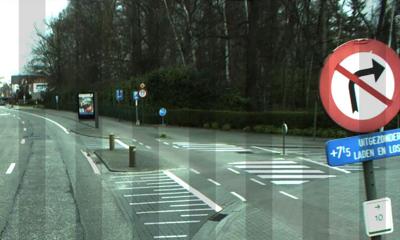}}\vspace{\vblank}\hfill
        {\includegraphics[width=\tempwidth]{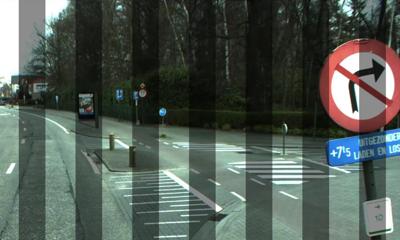}}\hfill
        {\includegraphics[width=\tempwidth]{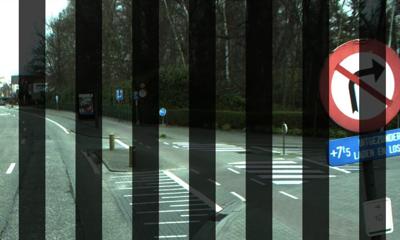}}\\
    \rowname{Snow}
        {\includegraphics[width=\tempwidth]{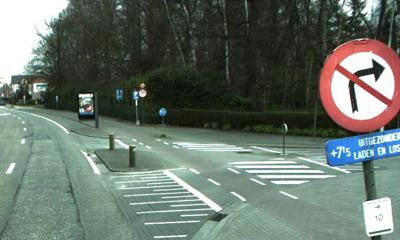}}\vspace{\vblank}\hfill
        {\includegraphics[width=\tempwidth]{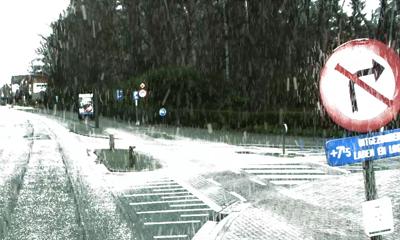}}\hfill
        {\includegraphics[width=\tempwidth]{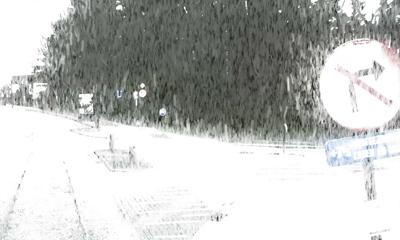}}\\
    \rowname{Haze}
        {\includegraphics[width=\tempwidth]{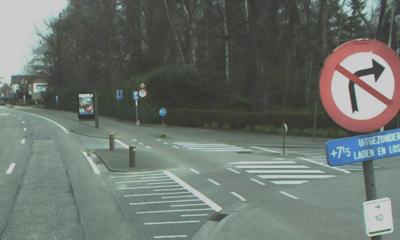}}\vspace{\vblank}\hfill
        {\includegraphics[width=\tempwidth]{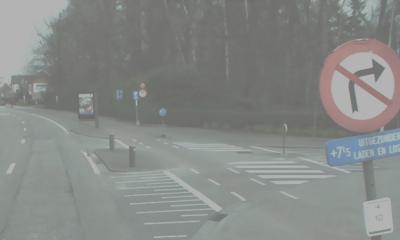}}\hfill
        {\includegraphics[width=\tempwidth]{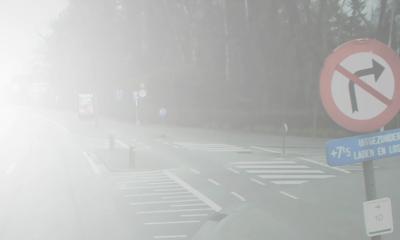}}\\

\caption{Sample scenes under challenging conditions with level 1, 3, and 5, respectively.}
\label{fig:challenges}

\end{figure}

\subsection{Challenging Conditions}
\label{subsec_chal_types}
\begin{figure*}[htbp!]
\centering
\begin{minipage}[b]{\linewidth}
  \centering
\includegraphics[width=0.9\linewidth]{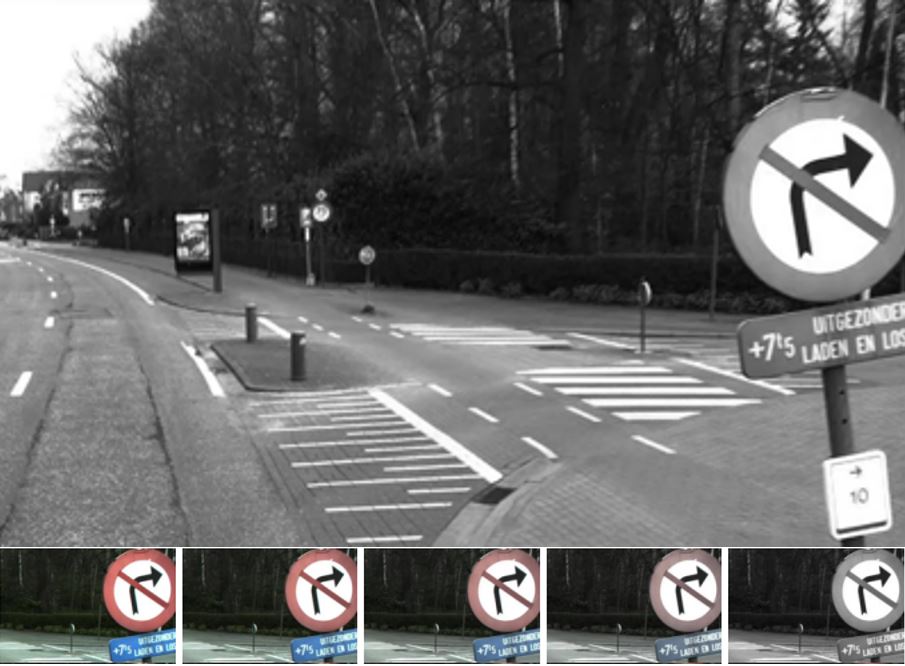}
  \vspace{0.1cm}
  \centerline{\footnotesize{(a) Decolorization condition in the CURE-TSD dataset. }}
  \vspace{0.1cm}
\end{minipage}
\begin{minipage}[b]{\linewidth}
  \centering
\includegraphics[width=0.9\linewidth]{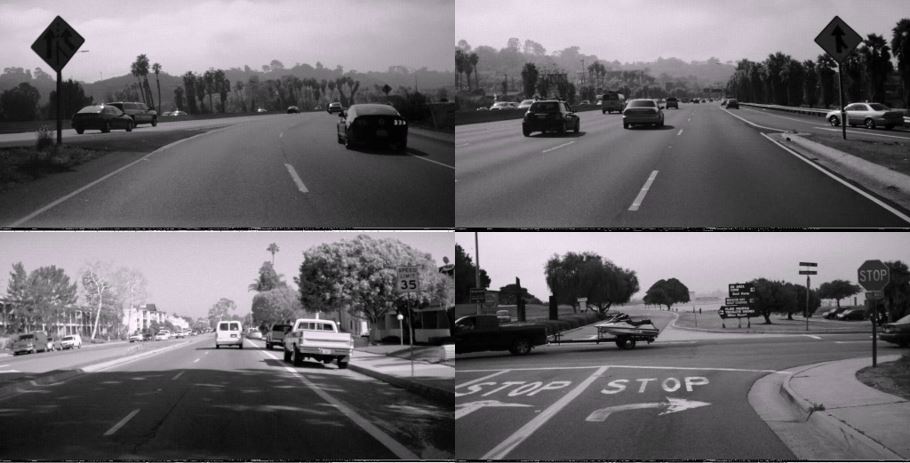}
  \vspace{0.1cm}
  \centerline{\footnotesize{(b) Decolorization condition in the LISA  dataset \cite{Mogelmose2012}. }}
  \vspace{0.1cm}
\end{minipage}
\caption{Decolorization condition in the CURE-TSD dataset and the LISA datset. Different challenge levels corresponding to the same scene are shown for the CURE-TSD dataset whereas different scene examples are shown for the LISA dataset.}
\label{fig:challenges_decolorization}
\end{figure*}

\begin{figure*}[htbp!]
\centering
\begin{minipage}[b]{\linewidth}
  \centering
\includegraphics[width=0.9\linewidth]{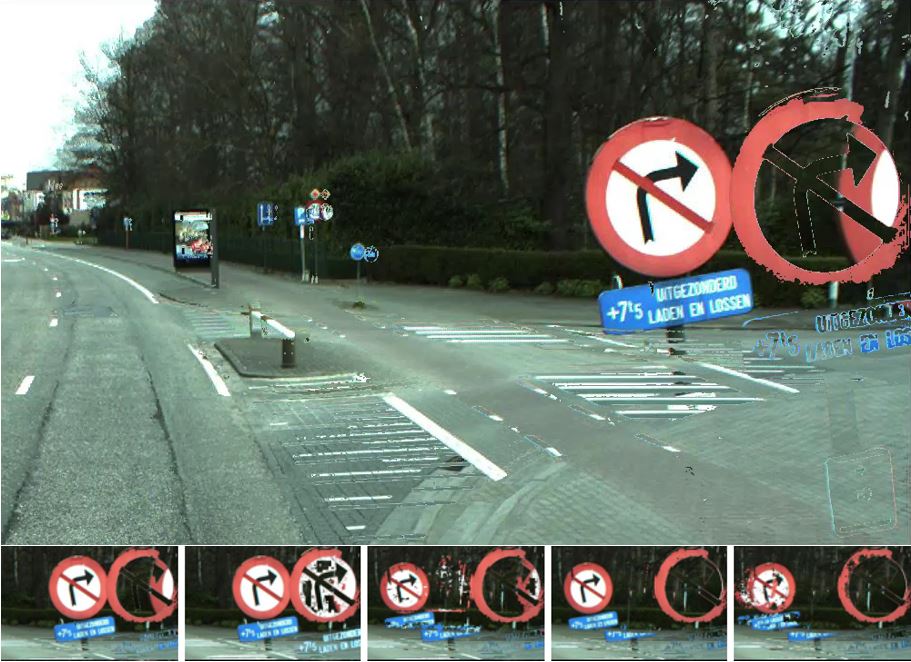}
  \vspace{0.1cm}
  \centerline{\footnotesize{(a) Codec error condition in the CURE-TSD dataset. }}
  \vspace{0.1cm}
\end{minipage}
\begin{minipage}[b]{\linewidth}
  \centering
\includegraphics[width=0.9\linewidth]{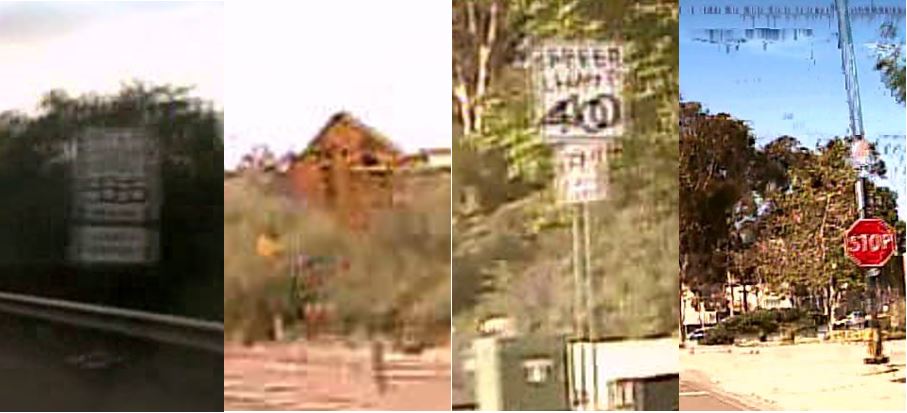}
  \vspace{0.1cm}
  \centerline{\footnotesize{(b) Codec error condition in the LISA dataset \cite{Mogelmose2012}. }}
  \vspace{0.1cm}
\end{minipage}
\caption{Codec error condition in the CURE-TSD dataset and the LISA datset. Different challenge levels corresponding to the same scene are shown for the CURE-TSD dataset whereas different scene examples are shown for the LISA dataset. Existing datasets are limited in terms of codec error severity. But we can still observe visual artifacts around traffic signs and displacements as shown in Fig.~\ref{fig:challenges_codecerror}(b).}
\label{fig:challenges_codecerror}
\end{figure*}

\begin{figure*}[htbp!]
\centering
\begin{minipage}[b]{\linewidth}
  \centering
\includegraphics[width=0.9\linewidth]{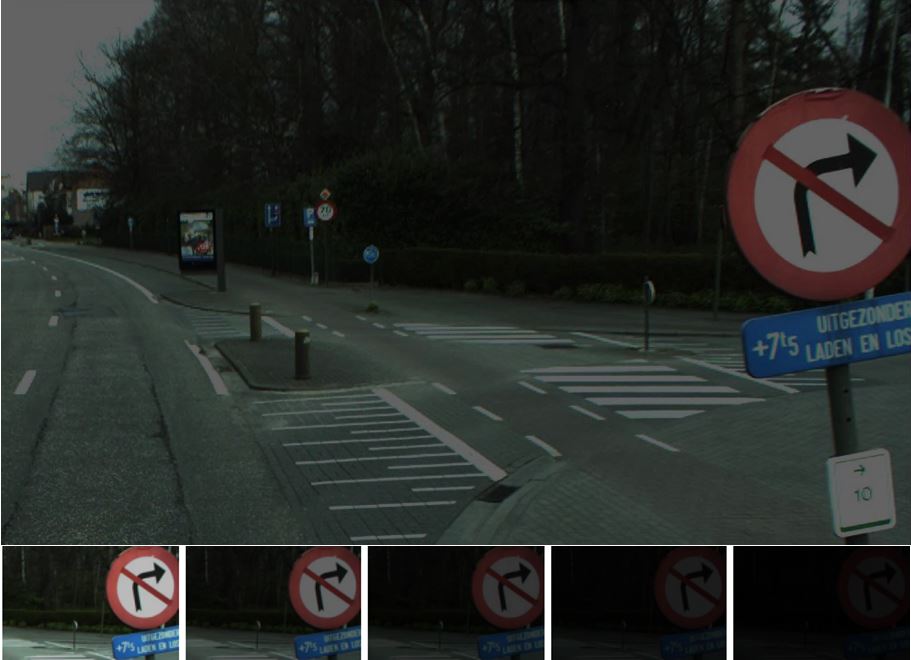}
  \vspace{0.1cm}
  \centerline{\footnotesize{(a) Darkening condition in the CURE-TSD dataset. }}
  \vspace{0.1cm}
\end{minipage}
\begin{minipage}[b]{\linewidth}
  \centering
\includegraphics[width=0.9\linewidth]{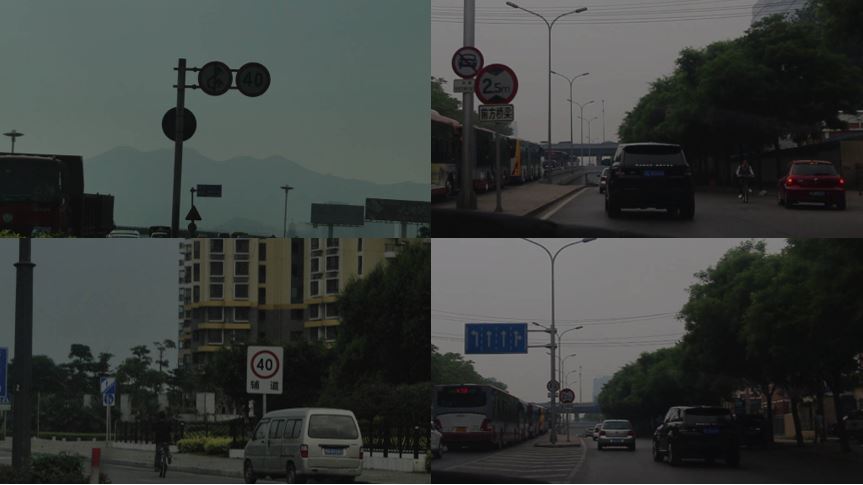}
  \vspace{0.1cm}
  \centerline{\footnotesize{(b) Darkening condition in the CCTSDB  dataset \cite{Zhang2017}. }}
  \vspace{0.1cm}
\end{minipage}
\caption{Darkening condition in the CURE-TSD dataset and the CCTSDB datset. Different challenge levels corresponding to the same scene are shown for the CURE-TSD dataset whereas different scene examples are shown for the CCTSDB dataset.}
\label{fig:challenges_darkening}
\end{figure*}

\begin{figure*}[htbp!]
\centering
\begin{minipage}[b]{\linewidth}
  \centering
\includegraphics[width=0.9\linewidth]{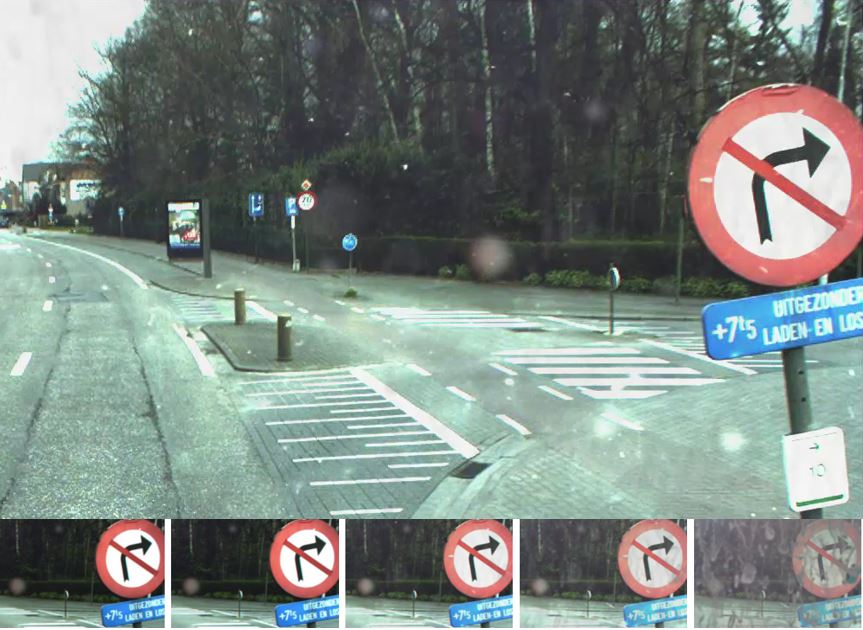}
  \vspace{0.1cm}
  \centerline{\footnotesize{(a) Dirty lens condition in the CURE-TSD dataset. }}
  \vspace{0.1cm}
\end{minipage}
\begin{minipage}[b]{\linewidth}
  \centering
\includegraphics[width=0.9\linewidth]{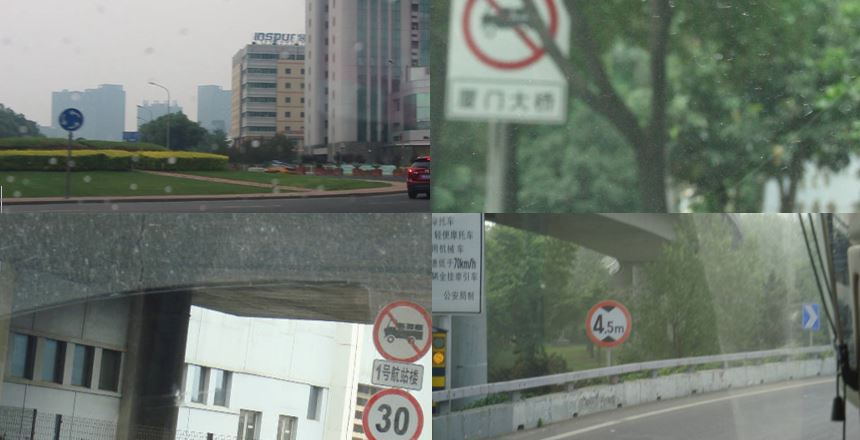}
  \vspace{0.1cm}
  \centerline{\footnotesize{(b) Dirty lens condition in the CCTSDB  dataset \cite{Zhang2017}. }}
  \vspace{0.1cm}
\end{minipage}
\caption{Dirty lens condition in the CURE-TSD dataset and the CCTSDB datset. Different challenge levels corresponding to the same scene are shown for the CURE-TSD dataset whereas different scene examples are shown for the CCTSDB dataset.}
\label{fig:challenges_dirtylens}
\end{figure*}

\begin{figure*}[htbp!]
\centering
\begin{minipage}[b]{\linewidth}
  \centering
\includegraphics[width=0.9\linewidth]{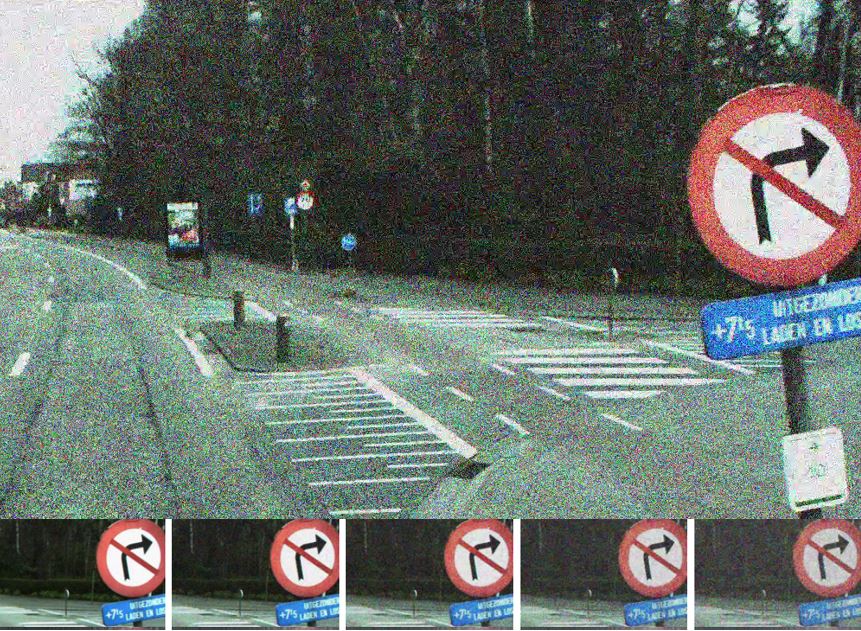}
  \vspace{0.1cm}
  \centerline{\footnotesize{(a) Noise condition in the CURE-TSD dataset. }}
  \vspace{0.1cm}
\end{minipage}
\begin{minipage}[b]{\linewidth}
  \centering
\includegraphics[width=0.9\linewidth]{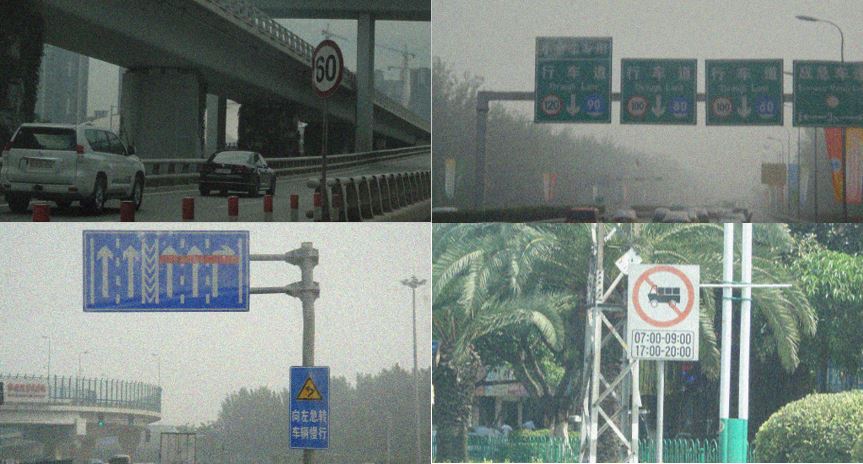}
  \vspace{0.1cm}
  \centerline{\footnotesize{(b) Noise condition in the CCTSDB  dataset \cite{Zhang2017}. }}
  \vspace{0.1cm}
\end{minipage}
\caption{Noise condition in the CURE-TSD dataset and the CCTSDB datset. Different challenge levels corresponding to the same scene are shown for the CURE-TSD dataset whereas different scene examples are shown for the the CCTSDB dataset.}
\label{fig:challenges_noise}
\end{figure*}

\begin{figure*}[htbp!]
\centering
\begin{minipage}[b]{\linewidth}
  \centering
\includegraphics[width=0.9\linewidth]{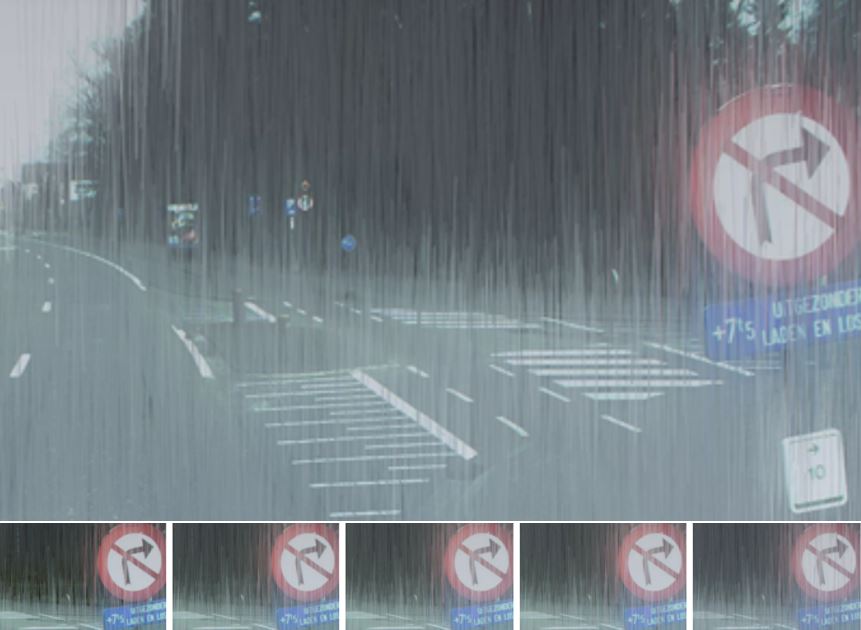}
  \vspace{0.1cm}
  \centerline{\footnotesize{(a) Rain condition in the CURE-TSD dataset. }}
  \vspace{0.1cm}
\end{minipage}
\begin{minipage}[b]{\linewidth}
  \centering
\includegraphics[width=0.9\linewidth]{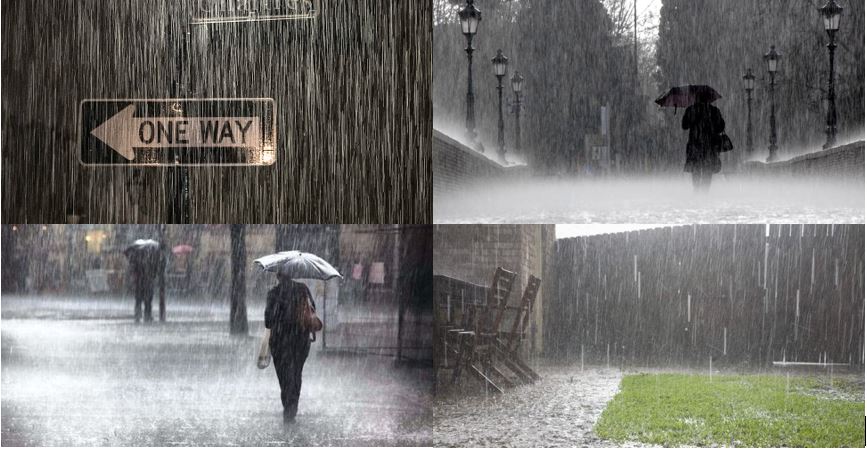}
  \vspace{0.1cm}
  \centerline{\footnotesize{(b) Rain condition in the DP Challenge \cite{dpchallenge} captured with a DSLR with 12-100mm f/4.0 lens (top left). Other images are obtained from image search. }}
  \vspace{0.1cm}
\end{minipage}
\caption{Rain condition in the CURE-TSD dataset, the DP challenge, and image search. Different challenge levels corresponding to the same scene are shown for the CURE-TSD dataset whereas different scene examples are shown for other images.}
\label{fig:challenges_rain}
\end{figure*}

\begin{figure*}[htbp!]
\centering
\begin{minipage}[b]{\linewidth}
  \centering
\includegraphics[width=0.9\linewidth]{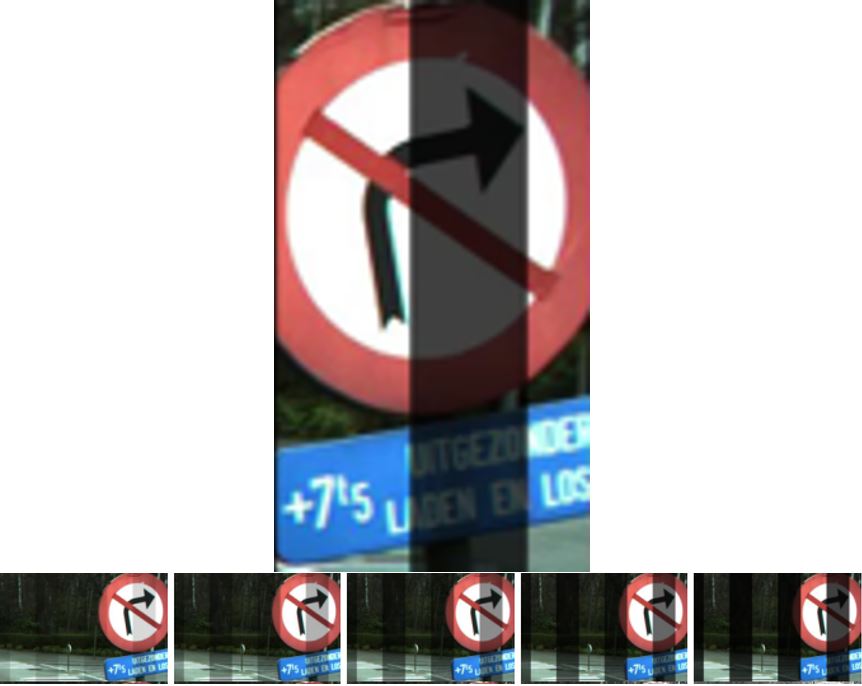}
  \vspace{0.1cm}
  \centerline{\footnotesize{(a) Shadow condition in the CURE-TSD dataset. }}
  \vspace{0.1cm}
\end{minipage}
\begin{minipage}[b]{\linewidth}
  \centering
\includegraphics[width=0.9\linewidth]{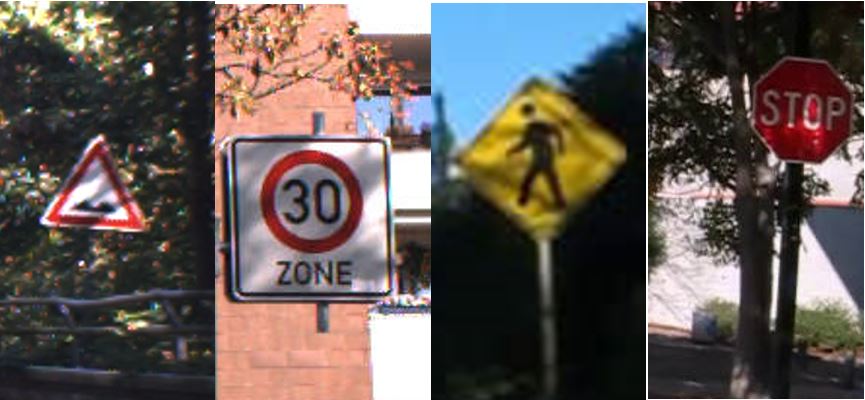}
  \vspace{0.1cm}
  \centerline{\footnotesize{(b) Shadow condition in the GTSDB dataset \cite{Houben2013} and the LISA dataset \cite{Mogelmose2012}. }}
  \vspace{0.1cm}
\end{minipage}
\caption{Shadow condition in the CURE-TSD dataset, the GTSDB dataset, and the LISA dataset. Different challenge levels corresponding to the same scene are shown for the CURE-TSD dataset whereas different scene examples are shown for other datasets. In the shadow challenge, traffic signs are partially occluded by dark shades. Images were cropped around the shadow region to observe its effect more clearly.}
\label{fig:challenges_shadow}
\end{figure*}

\begin{figure*}[htbp!]
\centering
\begin{minipage}[b]{\linewidth}
  \centering
\includegraphics[width=0.9\linewidth]{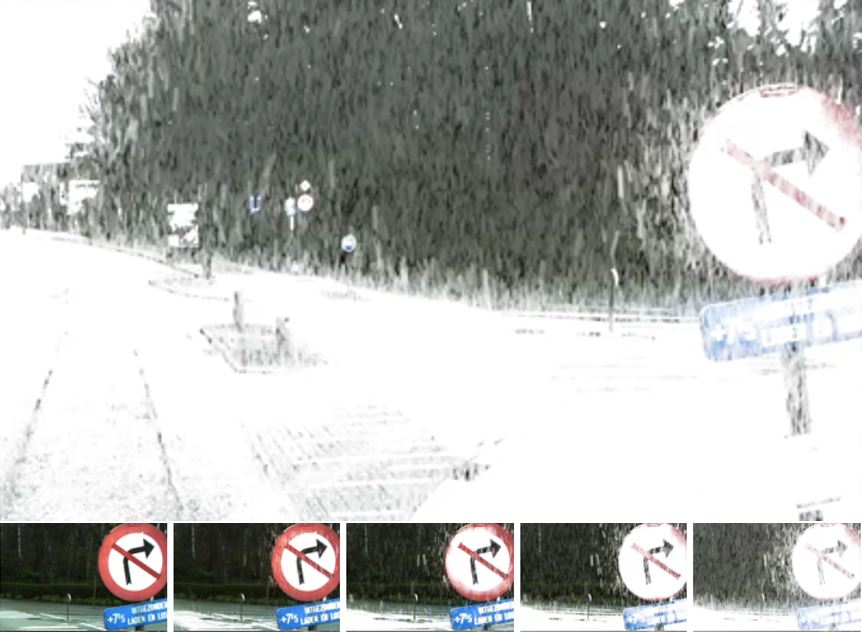}
  \vspace{0.1cm}
  \centerline{\footnotesize{(a) Snow condition in the CURE-TSD dataset. }}
  \vspace{0.1cm}
\end{minipage}
\begin{minipage}[b]{\linewidth}
  \centering
\includegraphics[width=0.9\linewidth]{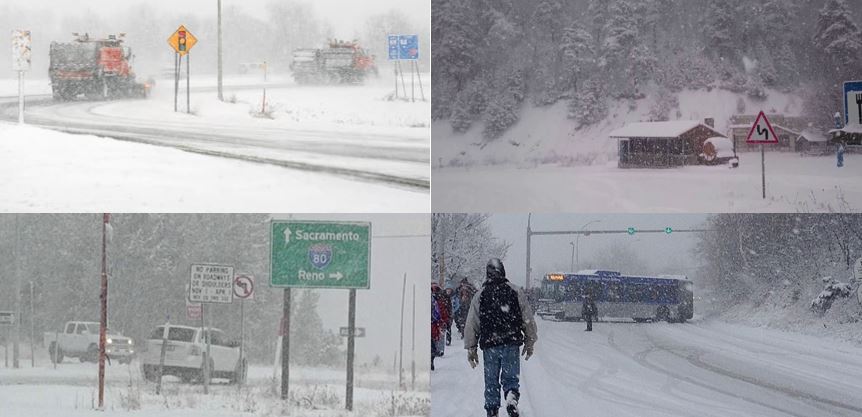}
  \vspace{0.1cm}
  \centerline{\footnotesize{(b) Snow condition images from image search. }}
  \vspace{0.1cm}
\end{minipage}
\caption{Snow condition in the CURE-TSD dataset and image search. Different challenge levels corresponding to the same scene are shown for the CURE-TSD dataset whereas different scene examples are shown for other images. In the snow category, external camera perspective is simulated in which camera has a direct view of the environment without a windshield as in TT-100K \cite{Zhu2016}.}
\label{fig:challenges_snow}
\end{figure*}

\begin{figure*}[htbp!]
\centering
\begin{minipage}[b]{\linewidth}
  \centering
\includegraphics[width=0.9\linewidth]{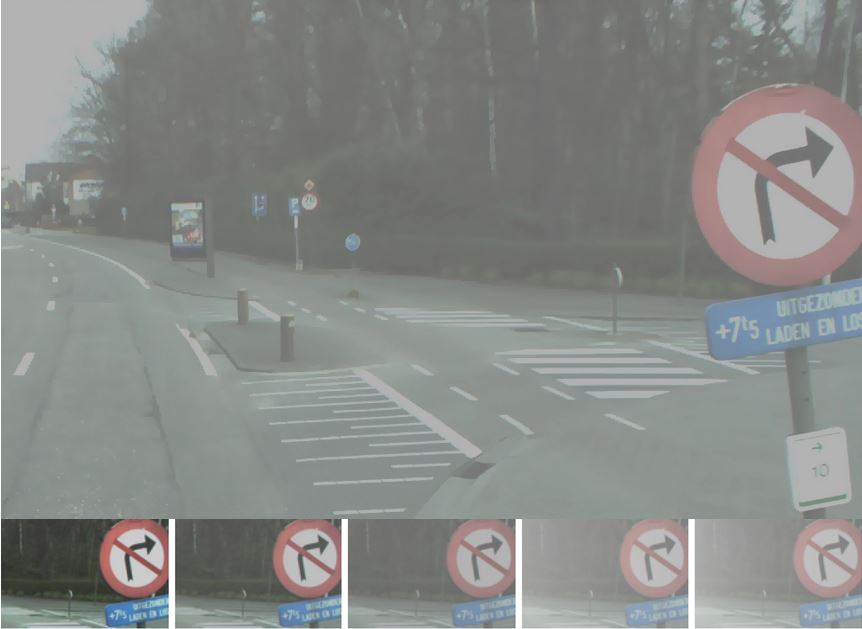}
  \vspace{0.1cm}
  \centerline{\footnotesize{(a) Haze condition in the CURE-TSD dataset. }}
  \vspace{0.1cm}
\end{minipage}
\begin{minipage}[b]{\linewidth}
  \centering
\includegraphics[width=0.9\linewidth]{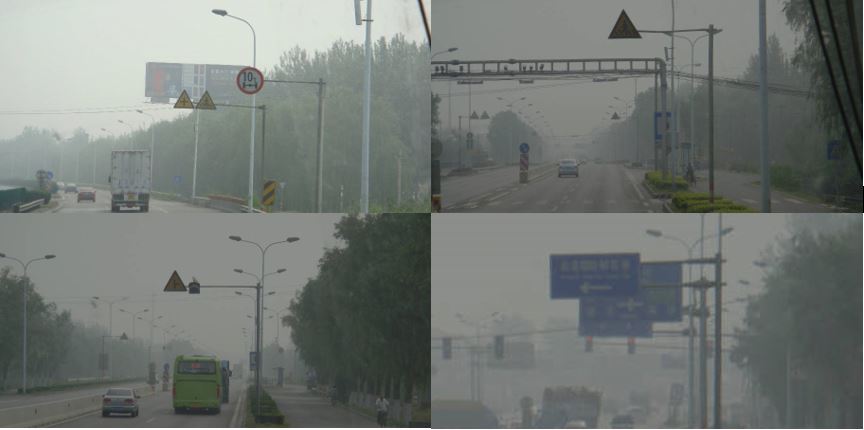}
  \vspace{0.1cm}
  \centerline{\footnotesize{(b) Haze condition in the CCTSDB dataset \cite{Zhang2017}. }}
  \vspace{0.1cm}
\end{minipage}
\caption{Haze condition in the CURE-TSD dataset and the CCTSDB dataset. Different challenge levels corresponding to the same scene are shown for the CURE-TSD dataset whereas different scene examples are shown the CCTSDB dataset.}
\label{fig:challenges_haze}
\end{figure*}

To emulate weather and vision system challenges, we utilized the Adobe(c) After Effects, which has been commonly used to synthesize visual effects and motion graphics at the post-production stage in previous studies \cite{Liu2013,Liu2014,Zhuang2017}. Control over environmental conditions and levels enable scaling up generated dataset size significantly to test algorithms for robustness. Challenge levels were adjusted for each challenge type separately through visual inspection.  We divided the challenges to five levels: level 1 does not affect the visibility of traffic signs from human perspective, level 2 affects the visibility of small and distant traffic signs, level 3 makes the visibility of small and distant traffic signs difficult, level 4 makes the visibility of small and distant traffic signs challenging, and level 5 makes the visibility of  small and distant traffic signs nearly impossible. We generated $12$ challenge types all of which correspond to conditions that can occur in real world. In Fig.~\ref{fig:challenges}, we show sample images from the \texttt{CURE-TSD} dataset that includes all the challenging conditions for level 1, 3, and 5. In Figs.~\ref{fig:challenges_decolorization}-\ref{fig:challenges_haze}, we include images of scenes with challenging conditions from the \texttt{CURE-TSD} dataset as well as other datasets with real-world acquisition to show the visual similarity between conditions.

\begin{figure*}[!h]
\setcounter{figure}{2}
\centering
\setlength{\tabcolsep}{0.2 em}
\scriptsize
\begin{tabular}{cccccccccccccc}
\includegraphics[width=0.06\linewidth,height=0.06\linewidth]{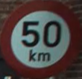} &
\includegraphics[width=0.06\linewidth,height=0.06\linewidth]{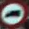} &
\includegraphics[width=0.06\linewidth,height=0.06\linewidth]{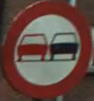} &
\includegraphics[width=0.06\linewidth,height=0.06\linewidth]{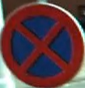} & \includegraphics[width=0.06\linewidth,height=0.06\linewidth]{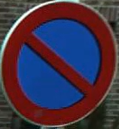} &
\includegraphics[width=0.06\linewidth,height=0.06\linewidth]{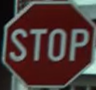} &
\includegraphics[width=0.06\linewidth,height=0.06\linewidth]{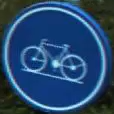} & \includegraphics[width=0.06\linewidth,height=0.06\linewidth]{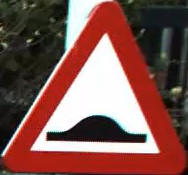} &
\includegraphics[width=0.06\linewidth,height=0.06\linewidth]{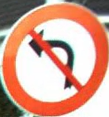} &
\includegraphics[width=0.06\linewidth,height=0.06\linewidth]{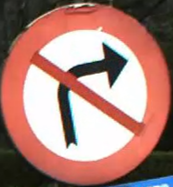} &
\includegraphics[width=0.06\linewidth,height=0.06\linewidth]{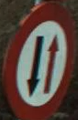} & \includegraphics[width=0.06\linewidth,height=0.06\linewidth]{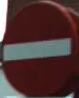} &
\includegraphics[width=0.06\linewidth,height=0.06\linewidth]{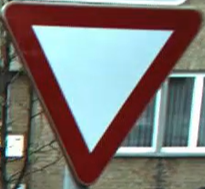} &
\includegraphics[width=0.06\linewidth,height=0.06\linewidth]{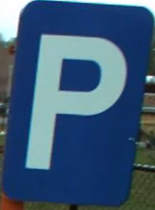} \\
\includegraphics[width=0.06\linewidth,height=0.06\linewidth]{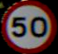} &
\includegraphics[width=0.06\linewidth,height=0.06\linewidth]{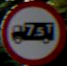} &
\includegraphics[width=0.06\linewidth,height=0.06\linewidth]{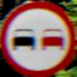} &
\includegraphics[width=0.06\linewidth,height=0.06\linewidth]{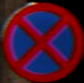} & \includegraphics[width=0.06\linewidth,height=0.06\linewidth]{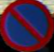} &
\includegraphics[width=0.06\linewidth,height=0.06\linewidth]{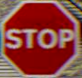} &
\includegraphics[width=0.06\linewidth,height=0.06\linewidth]{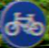} & \includegraphics[width=0.06\linewidth,height=0.06\linewidth]{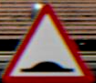} &
\includegraphics[width=0.06\linewidth,height=0.06\linewidth]{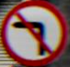} &
\includegraphics[width=0.06\linewidth,height=0.06\linewidth]{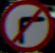} &
\includegraphics[width=0.06\linewidth,height=0.06\linewidth]{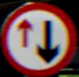} & \includegraphics[width=0.06\linewidth,height=0.06\linewidth]{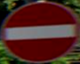} &
\includegraphics[width=0.06\linewidth,height=0.06\linewidth]{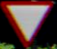} &
\includegraphics[width=0.06\linewidth,height=0.06\linewidth]{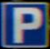} \\
speed & goods & no & no & no & \multirow{2}{*}{stop} & \multirow{2}{*}{bicycle} & \multirow{2}{*}{hump} & no & no & priority & no & \multirow{2}{*}{yield} & \multirow{2}{*}{parking} \\
limit & vehicles & overtaking & stopping & parking & & & & left & right & to & entry & & \\
\end{tabular}
\caption{Traffic sign samples from real-world ($1^{st}$ row) and synthesized  sequences ($2^{nd}$ row).}
\label{fig: sign_type}
\vspace{-4mm}
\end{figure*}

\begin{figure*}[!b]
\centering
\setlength{\tabcolsep}{0.2 em}
\scriptsize
\begin{tabular}{ccccccccccccc}
\includegraphics[width=0.06\linewidth,height=0.06\linewidth]{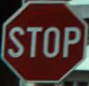} &
\includegraphics[width=0.06\linewidth,height=0.06\linewidth]{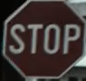} &
\includegraphics[width=0.06\linewidth,height=0.06\linewidth]{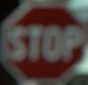} &
\includegraphics[width=0.06\linewidth,height=0.06\linewidth]{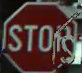} &
\includegraphics[width=0.06\linewidth,height=0.06\linewidth]{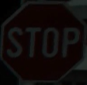} &
\includegraphics[width=0.06\linewidth,height=0.06\linewidth]{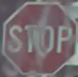} &
\includegraphics[width=0.06\linewidth,height=0.06\linewidth]{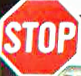} &
\includegraphics[width=0.06\linewidth,height=0.06\linewidth]{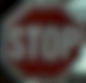} &
\includegraphics[width=0.06\linewidth,height=0.06\linewidth]{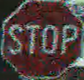} &
\includegraphics[width=0.06\linewidth,height=0.06\linewidth]{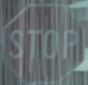} & \includegraphics[width=0.06\linewidth,height=0.06\linewidth]{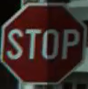} &
\includegraphics[width=0.06\linewidth,height=0.06\linewidth]{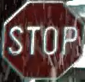} &
\includegraphics[width=0.06\linewidth,height=0.06\linewidth]{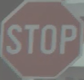} 
\\
\includegraphics[width=0.06\linewidth,height=0.06\linewidth]{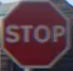} &
\includegraphics[width=0.06\linewidth,height=0.06\linewidth]{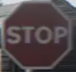} &
\includegraphics[width=0.06\linewidth,height=0.06\linewidth]{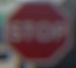} &
\includegraphics[width=0.06\linewidth,height=0.06\linewidth]{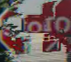} &
\includegraphics[width=0.06\linewidth,height=0.06\linewidth]{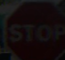} &
\includegraphics[width=0.06\linewidth,height=0.06\linewidth]{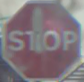} & 
\includegraphics[width=0.06\linewidth,height=0.06\linewidth]{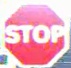} &
\includegraphics[width=0.06\linewidth,height=0.06\linewidth]{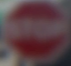} &
\includegraphics[width=0.06\linewidth,height=0.06\linewidth]{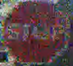} &
\includegraphics[width=0.06\linewidth,height=0.06\linewidth]{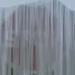} & \includegraphics[width=0.06\linewidth,height=0.06\linewidth]{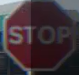} &
\includegraphics[width=0.06\linewidth,height=0.06\linewidth]{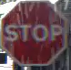} &
\includegraphics[width=0.06\linewidth,height=0.06\linewidth]{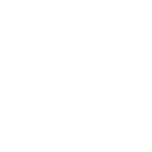} \\
No & Decolor- & Lens & Codec & \multirow{2}{*}{Darkening} & Dirty & \multirow{2}{*}{Exposure} & Gaussian & \multirow{2}{*}{Noise} & \multirow{2}{*}{Rain} & \multirow{2}{*}{Shadow} & \multirow{2}{*}{Snow} & \multirow{2}{*}{Haze} \\
Challenge & ization & Blur & Error & & Lens & & Blur & & & & & \\
\end{tabular}
\vspace{-2.0mm}
\caption{Stop sign samples under challenging conditions from real-world ($1^{st}$ row) and synthesized ($2^{nd}$ row) sequences.}
\label{fig: sign_distortion}
\end{figure*}

\texttt{Decolorization} tests the effect of color acquisition error, which was implemented with a color correction filter. \texttt{Lens Blur} and \texttt{Gaussian Blur} test the effect of dynamic scene acquisition, which was implemented with a smoothing operator. Unlike \texttt{Lens Blur}, \texttt{Gaussian Blur} is distributed in all directions, which results in less structured blurred scene. \texttt{Codec Error} tests the effect of encoder/decoder error, which was implemented with a time displacement filter. \texttt{Darkening} tests the effect of underexposure, which was implemented with an exposure filter. \texttt{Dirty Lens} tests the effect of occlusion because of dirt over camera lens, which was implemented by overlaying dirty lens images. \texttt{Exposure} tests the effect of overexposure in acquisition, which was implemented with an exposure filter. \texttt{Noise} tests the effect of acquisition noise, which was implemented using a noise filter. \texttt{Rain} tests the effect of occlusion due to rain, which was implemented using a gradient operator along with chroma filtering to obtain a blueish rain hue and a rain drop model.  \texttt{Shadow} tests the effect of non-uniform lighting due to shadow, which was implemented using a blind-shaped operator. \texttt{Snow} tests the effect of occlusion due to snow, which was implemented using a glow operator along with chroma filtering to obtain a white hue and a snow fall model. \texttt{Haze} tests the effect of occlusion due to haze, which was implemented using a radial gradient operator with partial opacity, a smoothing operator, an exposure operator, a brightness operator, and a contrast operator. Location of the operators were manually controlled to create a sense of depth.

\subsection{Traffic Signs}
Real-world sequences in the introduced dataset is based on the \texttt{BelgiumTS} dataset, which includes $62$ traffic sign types as summarized in Table \ref{tab_datasets}. While deciding on the type of traffic signs that should be annotated and included in the \texttt{CURE-TSD} dataset, we focused on two main criteria. First, not every sign type can be reasonably located in synthesized sequences. For example, we cannot reasonably include a level-crossing sign because there is no standard railroad feature in the utilized Unreal Engine 4 tool. Second, there are limited number of common signs between the package utilized in the generation of synthesized sequences and real-world sequences. Based on the aforementioned selection criteria, we provide $14$ types of traffic signs with annotations in all the sequences, which include speed limit, goods vehicles, no overtaking, no stopping, no parking, stop, bicycle, hump, no left, no right, priority to, no entry, yield, and parking, as shown in Fig.~\ref{fig: sign_type}. Each video sequence was processed with $12$ challenge types as described in Section~\ref{subsec_chal_types} except for haze in simulated sequences. To show the effect of challenging conditions over traffic signs, we show close up images of stop signs in Fig.~\ref{fig: sign_distortion}.

 \begin{figure}[h]
    \centering
    \includegraphics[width=\linewidth]{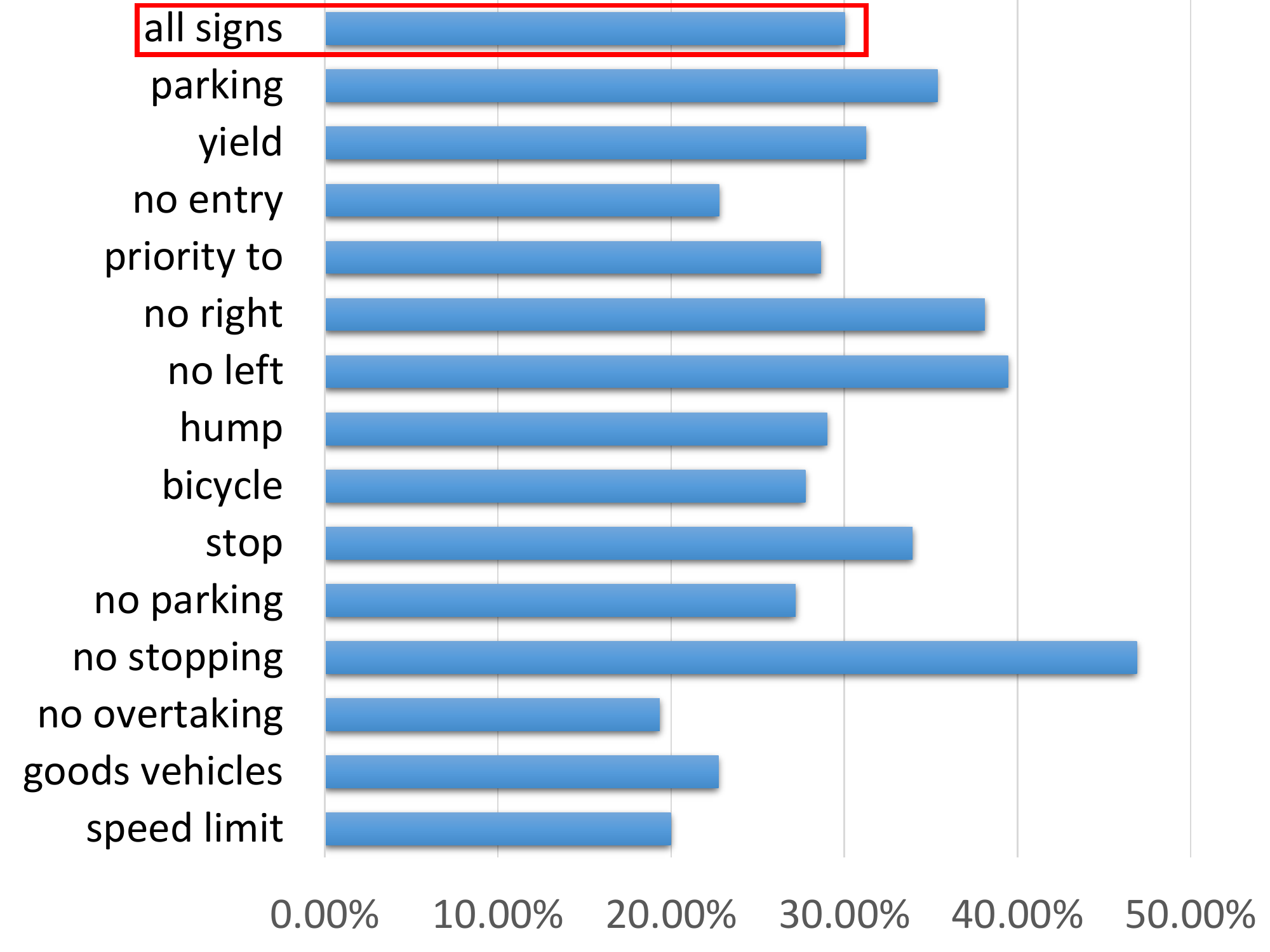}
    \caption{Ratio of signs in test set compared to overall dataset.}
    \label{fig:testing_ratio}
\end{figure}

 \subsection{Data Splitting}
To split the challenge-free sequences into training and test sets, we followed a commonly utilized splitting ratio $7:3$, which resulted in $68$ training videos and $30$ test videos. We split challenge videos based on the sequence they originate from so that $7:3$ ratio is maintained for challenge sequences as well. We started splitting video sequences based on the signs that appear the least and repeated the same procedure for the next less frequent sign type until each video sequence was classified. We obtained targeted split ratio for overall set with deviations up to $17\%$ in individual sign types as summarized in Fig.\ref{fig:testing_ratio}.

\subsection{Ground Truth Generation}
\begin{figure}
    \centering
    \includegraphics[width=0.95\linewidth]{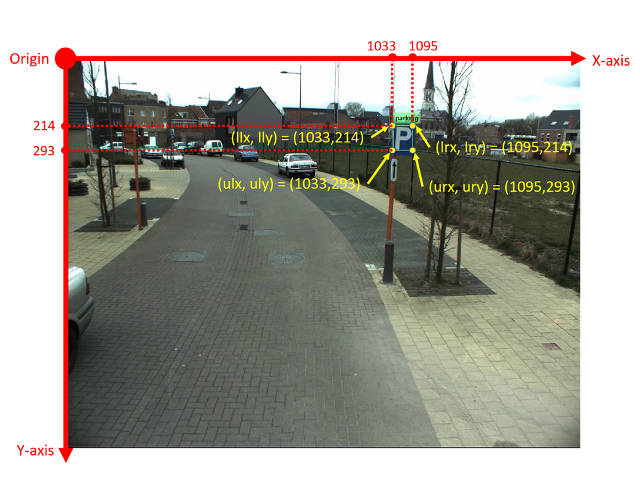}
    \caption{A sample annotated frame with the coordinate system.}
    \label{fig:coordinate}
\end{figure}

The annotations for BelgiumTS sequences were provided only for specific frames. However, our goal is to provide video sequences along with annotations for all frames to enable temporal information utilization. One option was to label those frames that lack annotations but that could have also led to inconsistency in the labelling process because of the inevitable differences across setups. Therefore, we annotated all frames in all BelgiumTS sequences using the Video Annotation Tool from Irvine, California (VATIC) \cite{vatic2}. We have utilized the pure \texttt{\href{https://dbolkensteyn.github.io/vatic.js/}{JavaScript version}}, which can directly run on a browser without requiring any specific installation.  A traffic sign was annotated if at least half of it was visible. A sample annotated frame with coordinate systems is shown in Fig.\ref{fig:coordinate}. 

\begin{figure}[h]
    \centering
    \includegraphics[width=\linewidth]{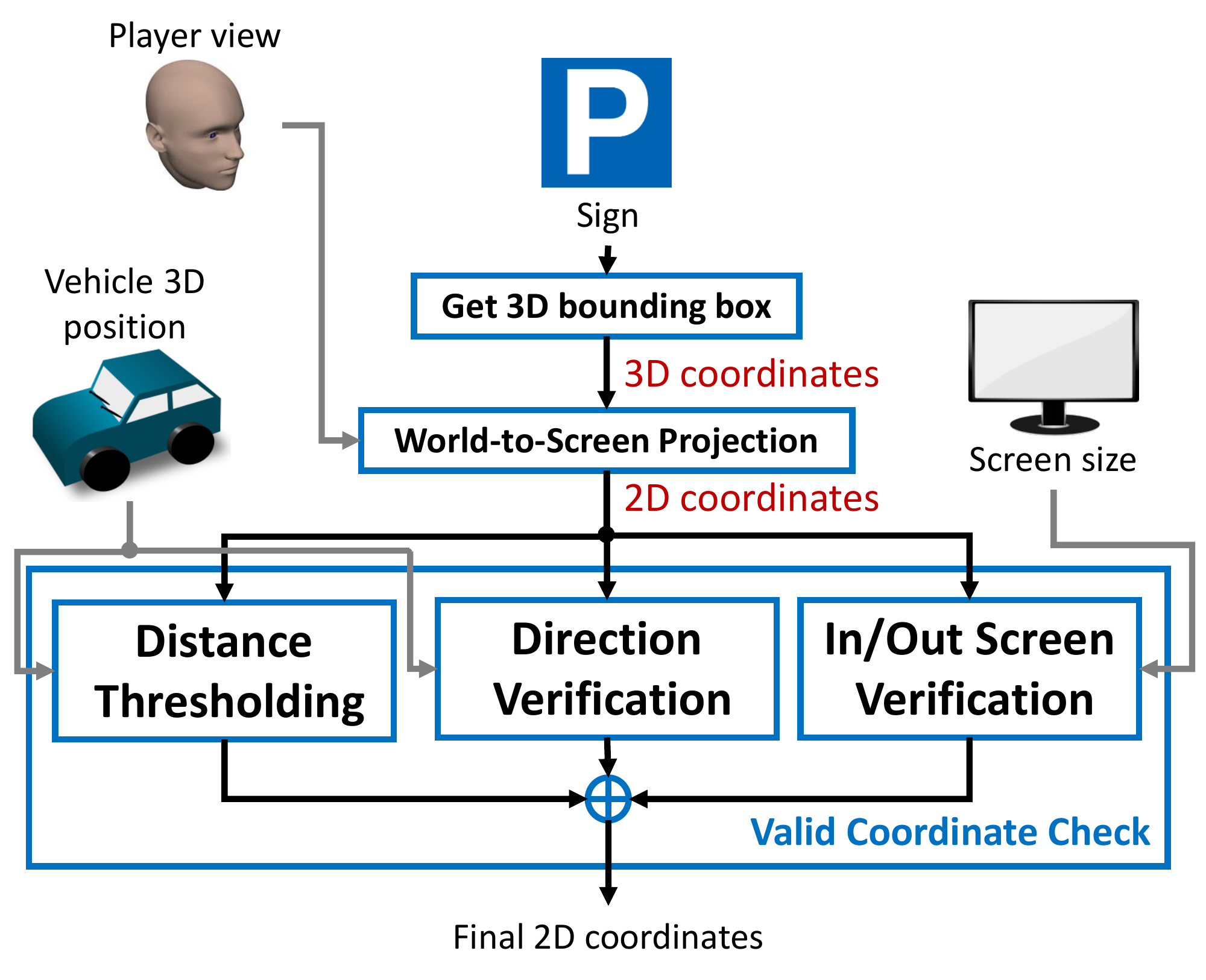}
    \caption{Annotation generation framework for synthesized sequences.}
    \label{fig:UE4}
\end{figure}

The annotations in synthesized sequences were obtained from the Unreal Engine 4 (\texttt{UE4}), whose framework is given in Fig.~\ref{fig:UE4} in which grey arrows indicate auxiliary information utilized for coordinate verification. Provided framework includes the operations that were performed to resolve four main issues in the annotation generation process. First, the built-in function for obtaining the bounding box coordinates in \texttt{UE4} outputs $3D$ world coordinates. We projected $3D$ world coordinates to $2D$ display coordinates. Second, \texttt{UE4} did not directly differentiate whether signs were inside the screen or not. To overcome the in/out of screen issue, we compared traffic sign coordinates with screen resolution and configured bounding boxes to cover the visible part. Third, \texttt{UE4} detected all bounding boxes in the virtual environment regardless of the relative distances. To eliminate the visibility issue, we set a distance threshold to filter out signs that were not perceivable. Fourth, \texttt{UE4} detected signs regardless of their direction. To resolve the direction issue, we calculated the angle between the direction of the vehicle and the direction of the front side of the sign and excluded the traffic signs with less than $90$ degrees. 

\section{Benchmark Algorithms}
We introduced the \texttt{CURE-TSD} dataset in Video and Image Processing (VIP) Cup 2017 competition whose theme was \emph{Traffic Sign Detection under Challenging Conditions}. We hosted the competition as part of the IEEE International Conference on Image Processing (ICIP) 2017, which was sponsored by IEEE Signal Processing Society (SPS) and supported by the SPS Multimedia and Signal Processing (MMSP) Technical Committee. VIP Cup 2017 started with more than 250 requests from 147 parties from all around the world and $19$ of these teams with a total of 80 members were able to attend the competition by satisfying the participation requirements. In this paper, we analyze the algorithms of top four performing teams and refer to the algorithms with the name of the teams as \texttt{Neurons}, \texttt{PolyUTS}, \texttt{IIP}, and \texttt{Markovians}.

\subsection{Algorithm Neurons}

\vspace{2 mm}
\noindent{\bf Preprocessing:}~A customized preprocessing is performed over video frames depending on the challenge type to enhance traffic sign features. For video sequences with decolorization, darkening, exposure, shadow, snow, and haze, contrast-limited adaptive histogram equalization (CLAHE) was used. First, frames are converted from RGB domain to HSV domain. Then, different configurations of CLAHE are performed over the saturation and value channels depending on the challenge type. For video sequences with rain conditions, a ResNet type of CNN \cite{He2015} is used to eliminate the effect of rain as in \cite{Fu2017}, which includes consecutive blocks that include convolution, batch normalization, and ReLU along with skip connections. 

\begin{table}[htbp!]
\caption{Challenge classification network of Neurons.}
\label{tab:neurons_challenge_classification}
\centering
{\renewcommand{\arraystretch}{1.2}
\begin{tabular}{ccc}
\hline
 \bf Layer type   &\bf Kernel size \& parameters  \\ \hline
 input &3x309x407  \\ 
 convolution &3x3  \\ 
 convolution  &3x3  \\ 
 max pooling  &4x4  \\ 
 convolution &3x3  \\ 
 convolution  &3x3  \\ 
 max pooling &2x2  \\ 
 convolution &3x3  \\ 
 convolution &3x3  \\ 
 max pooling &2x2  \\ 
 convolution &3x3  \\ 
 convolution &3x3  \\ 
 max pooling &2x2  \\ 
 vectorization  &-  \\ 
 fully connected with ReLU &500 \\ 
 fully connected with Softmax  &12  \\ \hline
\end{tabular}
}
\end{table}

\vspace{2 mm}
\noindent{\bf Challenge classification:}~\texttt{Neurons} detected challenge types with a CNN-based architecture similar to VGG-16 \cite{Simonyan2014}. There are four main blocks with 2 convolutional layers, ReLU and a max-polling layer. After the fourth layer, feature maps are vectorized and mapped to challenge types with two fully connected layers. Details of the challenge classification architecture are given in Table \ref{tab:neurons_challenge_classification}.

\begin{table}[htbp!]
\caption{Traffic sign classification network of Neurons.}
\label{tab:neurons_sign_classification}
\centering
{\renewcommand{\arraystretch}{1.2}
\begin{tabular}{cc}
\hline
 \bf Layer type &\bf Kernel size \& parameters  \\ \hline
 input&3x32x32  \\ 
 convolution  &3x3  \\ 
 convolution  &3x3  \\ 
 max pooling  &2x2  \\ 
 drop out & 0.25  \\ 
 convolution  &3x3  \\ 
 convolution  &3x3  \\ 
 max pooling  &2x2  \\ 
 drop out & 0.25  \\ 
 vectorization  &-  \\ 
 fully connected with ReLU &1024 \\ 
 fully connected with Softmax &14  \\ \hline
\end{tabular}
}
\end{table}

\vspace{2 mm}
\noindent{\bf Detection:}~ Traffic signs are localized with challenge-specific U-Net architecture \cite{Ronneberger2015} and detected signs are classified with another CNN architecture. Input images are resized to quarter of the original image resolution to reduce the number of parameters in the network. Localization architecture is based on downsampling stage and upsampling stage. In the downsampling stage, there are consecutive blocks that include convolution, batch normalization, ReLU, and max-pooling. Number of feature maps are doubled after each block. In the upsampling stage, input of each block is concatenated with a cropped feature map from the corresponding downsampling block. At the end of upsampling stage, convolutional layers map the features to binary classes as background or traffic sign. Localized regions are classified with a shallow CNN to recognize traffic sign types as summarized in Table \ref{tab:neurons_sign_classification}. There are two main blocks in the classification architecture, which includes two consecutive convolutinal layers, a ReLU layer, a max-pooling layer followed by a dropout layer. Feature vectors are mapped to traffic sign types with two fully connected layers.

\subsection{Algorithm PolyUTS}

\vspace{2 mm}
\noindent{\bf Detection:}~Input images are resized to 640x480 before being fed to the detection architecture. \texttt{PolyUTS} utilized pre-trained GoogLeNet \cite{Szegedy2014} as feature extractor and features at the end of Inception 5B layer are fed into regression layers to obtain coordinates with confidence scores. Proposed detection-by-regression approach was inspired by state-of-the-art YOLO architecture \cite{Redmon2015}. In the YOLO architecture, feature vectors are concatenated and then fed to fully connected layers. However, \texttt{PolyUTS} passes the feature vectors one by one to the fully connected layers without concatenation to maintain the spatial information. Pretrained network outputs feature maps of size 20x15x1024. Then, each 20x15 feature map is fed to the regression stage that is based on two fully connected layers. Duplicated regions are eliminated with non-maximum suppression and a CNN-based architecture is used to classify region of interests into $14$ classes of road signs and $1$ class of non-sign region. Details of the traffic sign classification architecture are given in Table~\ref{tab:polyuts_sign_classification}.

\begin{table}[htbp!]
\caption{Traffic sign classification network of PolyUTS.}
\label{tab:polyuts_sign_classification}
\centering
{\renewcommand{\arraystretch}{1.2}
\begin{tabular}{cc}
\hline
\bf Layer type  &\bf Kernel size \& parameters  \\ \hline
input  &3x48x48  \\ 
convolution with ReLU and batch norm &7x7x100  \\ 
max pooling &3x3 with stride 2  \\ 
convolution with ReLU and batch norm  &4x4x150  \\ 
max pooling &3x3 with stride 2  \\ 
convolution with ReLU and batch norm  &4x4x250  \\ 
max pooling &3x3 with stride 2  \\ 
vectorization   &-  \\ 
fully connected with ReLU and batch norm  &300 \\ 
fully connected with Softmax  &15  \\ \hline
\end{tabular}
}
\end{table}

\subsection{Algorithm IIP}
\vspace{2 mm}
\noindent{\bf Detection:}~\texttt{IIP} utilized a Faster R-CNN architecture \cite{Ren2015} to detect traffic signs in the \texttt{CURE-TSD} dataset. Faster R-CNN is based on a region proposal network (RPN) and an object detector. RPN receives an image without size constraints and outputs a set of rectangular regions with an objectness score. Region proposals correspond to \emph{attention} mechanisms \cite{Chorowski2015}, which tells the model where to pay attention. Zeiler and Fergus (ZF) \cite{Zeiler2013} model is used to minimize the overall computation required for region proposal and object detection. Specifically, there are 5 sharable convolutional layers whose outputs are used for region proposals as well as object detection. A small network is slid over the last shared convolutional layer and at each sliding window, multiple region proposals are predicted with various scale and aspect ratios. Training of Faster R-CNN is based on four steps. First, RPN is initialized with an ImageNet pretrained model and then fine-tuned end-to-end. Second, object detection network Fast R-CNN \cite{Girshick2015} is also initialized by the ImageNet pretrained model and trained with the region proposals genrated by RPN. Third, Fast R-CNN network is used to initialize the RPN training by fixing the shared convolutional layers and fine-tuning the remaining layers. Fourth, shared convolutional layers are kept fixed while fine-tuning the remaining layers of Fast R-CNN. Non-maximum suppression is used to merge highly overlapped bounding boxes and proposed regions are filtered with a fixed threshold to determine final results.

\begin{table}[htbp!]
\caption{Shared convolutional layers between region proposal network and object detector in Markovians.}
\label{tab:markovians_sign_classification}
\centering
{\renewcommand{\arraystretch}{1.2}
\begin{tabular}{cc}
\hline
 \bf Layer type   &\bf Kernel size \& parameters  \\ \hline
input  &224x224x3  \\ 
convolution with ReLU  &96x7x7 with a stride of 2  \\ 
max pooling and contrast norm&3x3 with stride a stride of 2  \\ 

convolution with ReLU  &256x5x5 with a stride of 2  \\ 
max pooling and contrast norm&3x3 with stride a stride of 2  \\ 

convolution with ReLU  &384x3x3 with a stride of 1  \\ 

convolution with ReLU  &384x3x3 with a stride of 1  \\ 

convolution with ReLU  &256x3x3 with a stride of 1  \\ 
max pooling &3x3 with stride a stride of 2  \\ \hline
\end{tabular}
}
\end{table}

\subsection{Algorithm Markovians}
\vspace{2 mm}
\noindent{\bf Challenge Classification:}~\texttt{Markovians} classified videos based on challenge types with a Recurrent CNN \cite{Liang2015}. First layer of recurrent CNN is a standard feed-forward convolutinal layer followed by max pooling. The following four layers are based on recurrent convolutinal layers (RCL) with a max pooling layer in the middle. Adjacent RCL layers are connected with feed-forward connections. Pooling layers have a size of 3x3 with a stride of 2. There is a global max pooling layer at the end of last RCL layer for every feature map, which results in the feature vector representing the input image. Finally, a softmax classifier is used to map feature vectors to challenge categories. Estiamted challenge classes are used to determine sign tracking mechanism.

\begin{table}[htbp!]
\caption{Traffic sign classification network of Markovians.}
\label{tab:markovians_sign_classification}
\centering
{\renewcommand{\arraystretch}{1.2}
\begin{tabular}{cc}
\hline
 \bf Layer type   &\bf Kernel size \& parameters  \\ \hline
 input  &64x64x3  \\ 
convolution with ReLU  &32x3x3  \\ 
max pooling&2x2  \\ 
drop out&0.25  \\ 
convolution with ReLU  &64x3x3  \\ 
convolution with ReLU  &64x3x3  \\ 
max pooling&2x2  \\ 
drop out&0.25  \\ 
vectorization&-  \\ 
fully connected with ReLU   &512 \\ 
drop out &0.5  \\ 
fully connected with Sigmoid   &24 \\ \hline
\end{tabular}
}
\end{table}

\vspace{2 mm}
\noindent{\bf Detection:}~Input images are resized to 640x480 before
being fed to the detection architecture. Similar to \texttt{IIP}, \texttt{Markovians} localized traffic signs with Faster R-CNN along with non-maximum suppression to merge highly overlapped estimations. A custom CNN-based approach is utilized to classify traffic signs whose details are provided in Table~\ref{tab:markovians_sign_classification}.

\vspace{2 mm}
\noindent{\bf Tracking:}~In case of dirty lens and shadow, Kalman filtering is used to track signs. Estimated sign locations are considered valid if the distance between current measurement and prediction is less than 50 pixels. In case of other challenging conditions, traffic signs are tracked with Lucas-Kanade algorithm \cite{Lucas1981}, which is a commonly used differential method for optical flow computation that is based on the assumption that flow between consecutive frames in a local neighborhood is constant. 

 \begin{center}
\begin{table*}[!b]
\begin{center}
\caption{Characteristics and performance of top performing algorithms. }
\label{tab:results}
{\renewcommand{\arraystretch}{1.2}
 \begin{tabular}{>{\centering\arraybackslash}p{0.9cm} c| >{\centering\arraybackslash}p{0.9cm} >{\centering\arraybackslash}p{3.0cm} >{\centering\arraybackslash}p{3.6cm}  >{\centering\arraybackslash}p{1.6cm} |c c c c} 
\hline
\multicolumn{2}{c|}{Team} &  \multicolumn{4}{c|}{Method}  &  \multicolumn{4}{c}{Detection Performance} \\ \hline
 Name & ID & \begin{tabular}[c]{@{}c@{}}Challenge \\classification \end{tabular} &\begin{tabular}[c]{@{}c@{}}Custom \\preprocessing \end{tabular} & Detection &Tracking & $Precision$ & $Recall$ & $F_{0.5}$ & $F_{2}$ \\ [0.5ex] 
 \hline
 IIP           & 24,873  &  - &- &Faster R-CNN&-  & 0.409 & 0.249 & 0.363 & 0.270 \\ 

 Markovians     & 24,839 &\begin{tabular}[c]{@{}c@{}}Recurrent \\CNN  \end{tabular}  &- & \begin{tabular}[c]{@{}c@{}}Faster R-CNN-based localization \\ Custom CNN-based recognition \end{tabular}& \begin{tabular}[c]{@{}c@{}}Kalman filtering, \\Lucas-Kanade \end{tabular} & 0.509 & 0.250 & 0.422 & 0.279 \\ 

 Neurons        & 24,861 &\begin{tabular}[c]{@{}c@{}}CNN \\($\sim$VGG)  \end{tabular}   & \begin{tabular}[c]{@{}c@{}}Color space transformation \\ histogram equalization (CLAHE) \\ResNet-based denoising \end{tabular}& \begin{tabular}[c]{@{}c@{}}U-Net-based localization, \\Custom CNN-based recognition \end{tabular} &-    & 0.550 & 0.320 & 0.481 & 0.349 \\ 

 PolyUTS        & 24,860 & - &- & \begin{tabular}[c]{@{}c@{}}CNN-based GoogLeNet-based localization \\Custom CNN-based recognition \end{tabular}       &- & 0.504 & 0.284 & 0.436 &0.311 
\\ \hline
\end{tabular}
}
\end{center}
\end{table*}

\end{center}

\subsection{Overview of the Benchmark Algorithms}    
We summarize the main characteristics of top performing traffic sign detection algorithms in the VIP Cup 2017 in Table~\ref{tab:results}. All of these algorithms are based on state-of-the-art architectures including VGG-16 \cite{Simonyan2014}, GoogLeNet \cite{Szegedy2014}, U-net \cite{Ronneberger2015}, ResNet \cite{He2015}, YOLO \cite{Redmon2015}, Recurrent CNN \cite{Liang2015}, Fast R-CNN \cite{Girshick2015}, and Faster R-CNN \cite{Ren2015}. There are four main differences between finalist algorithms and existing algorithms. First, video sequences were classified based on their challenge types by teams \texttt{Markovians} and \texttt{Neurons}. Second, video sequences were preprocessed based on challenge types to eliminate the effect of challenges by team \texttt{Neurons}. Third, team \texttt{Markovians} utilized temporal information in their algorithm, which is overlooked by majority of state-of-the-art architectures in the literature. Fourth, tracking mechanism is selected based on the classified challenge category by team \texttt{Markovians}.

\vspace{4mm}
\section{Performance Benchmark for CURE-TSD Dataset}
\subsection{Validation Metrics}
We validated competing algorithms based on the detection results over test sequences. Specifically, we computed $Precision$, $Recall$, and combination of these metrics ($F$-scores) to rank the performance of competing teams. $Precision$ is calculated as
\begin{equation}
Precision = \frac{TP}{TP+FP},
\end{equation}
where $TP$ is the total number of true positive detections across all test sequences, and $FP$ is the total number of false positive detections across all test sequences. A true positive is obtained when the overlap area between the ground truth box and the estimated box divided by the smaller box area exceeds $50\%$. When normalized overlap is less than $50\%$, estimate leads to a false positive.  $Recall$ is calculated as
\begin{equation}
Recall = \frac{TP}{TP+FN},
\end{equation}
where $TP$ is the total number of true positive detections across all test sequences and $FN$ corresponds to the total number of undetected traffic signs. Furthermore, we use $F$-scores to combine $Precision$ and $Recall$, which is formulated as
\begin{equation}
F_{\beta} = (1+{\beta}^2) \frac{Precision \cdot Recall}{{\beta}^2 Precision+Recall},
\end{equation}
where $\beta$ is a parameter used to adjust relative significance of $Precision$ and $Recall$. In our validation, we set $\beta$ to $2$ and $0.5$, which are commonly utilized in detection tasks.

\begin{figure*}[htbp!]
\centering
\begin{minipage}[b]{0.45\linewidth}
  \centering
\includegraphics[width=\linewidth, trim= 40mm 85mm 40mm 85mm]{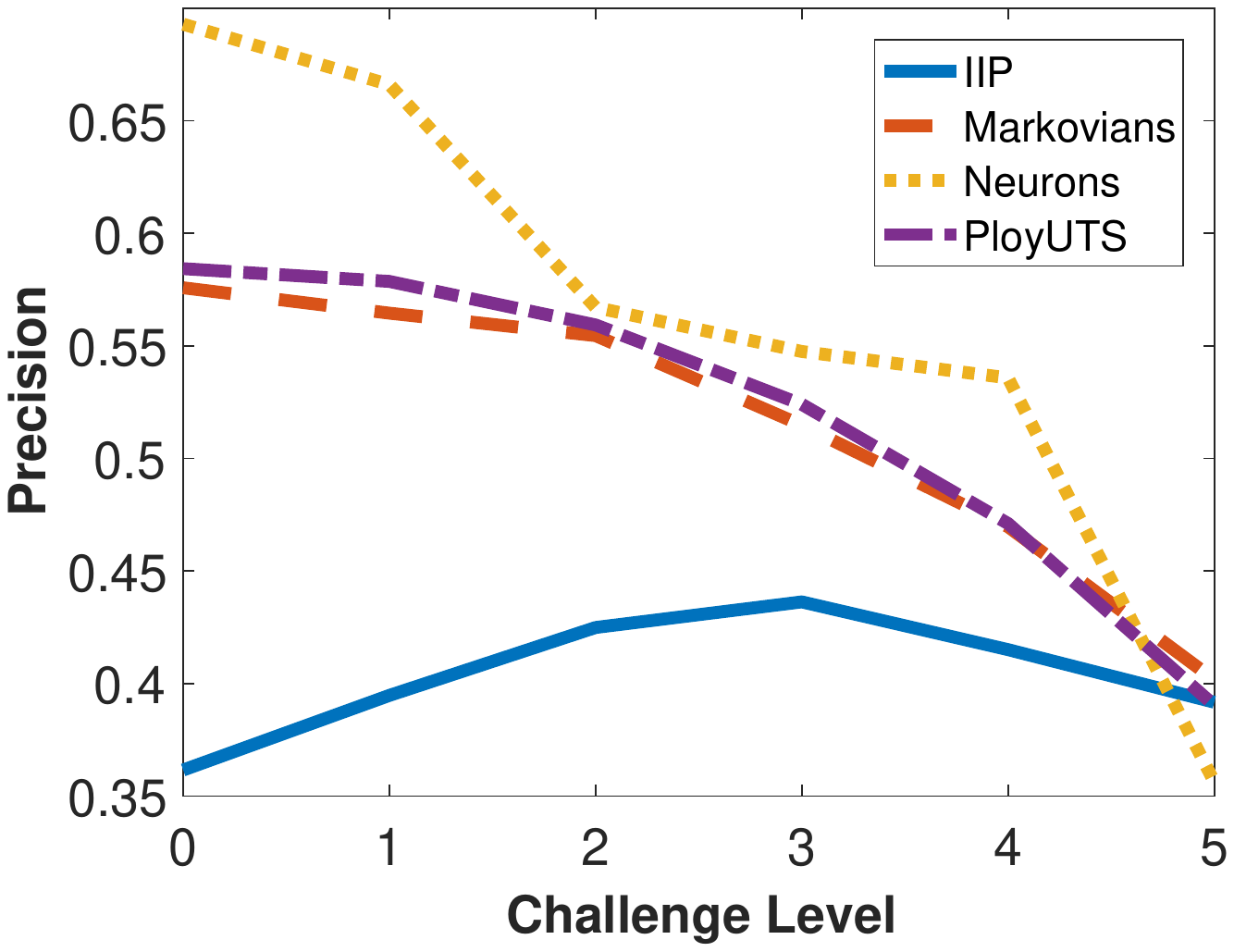}
  \vspace{0.01cm}
  \centerline{\footnotesize{(a)Precision}}
\end{minipage}
\begin{minipage}[b]{0.45\linewidth}
  \centering
\includegraphics[width=\linewidth, trim= 40mm 85mm 40mm 85mm]{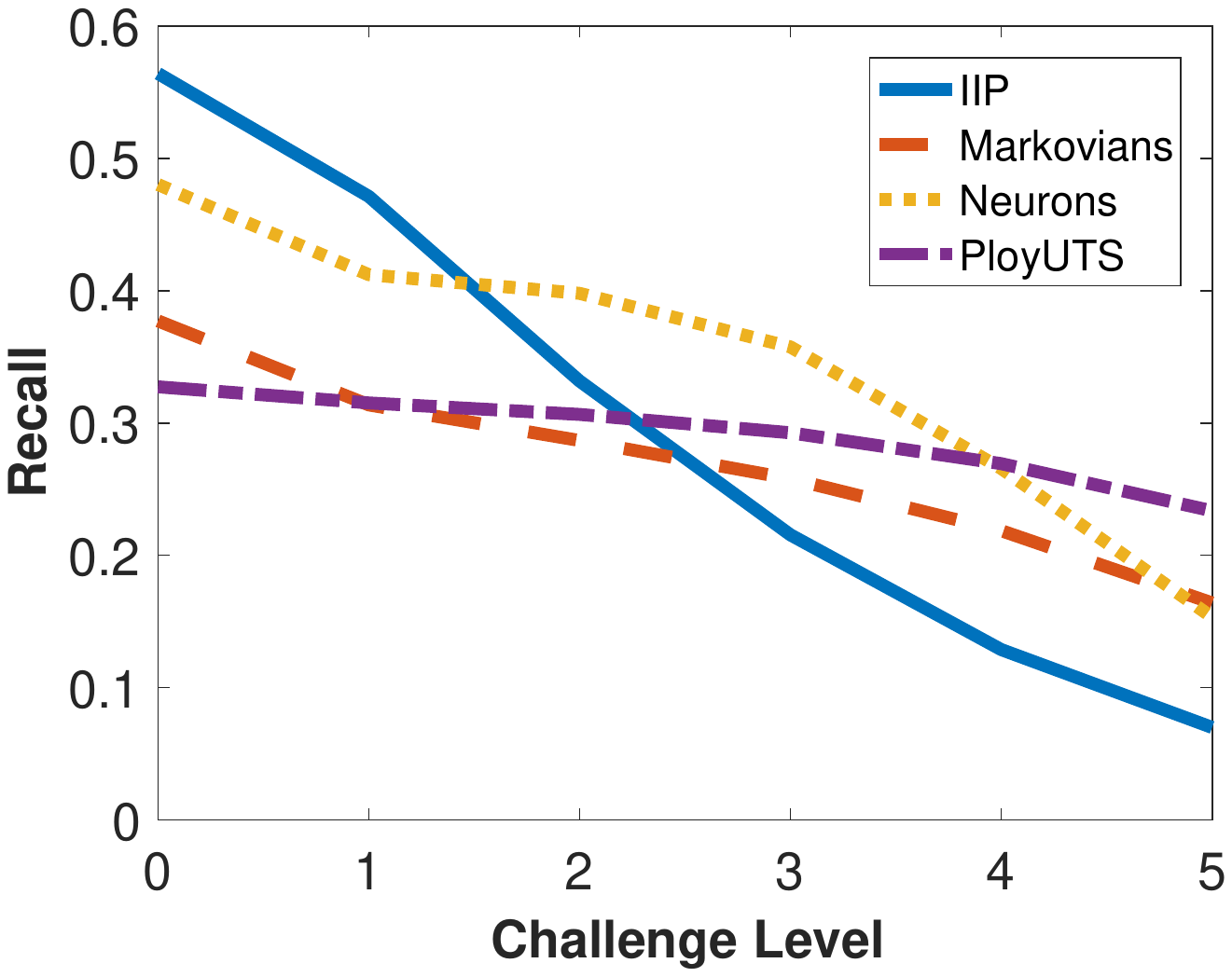}
  \vspace{0.01cm}
  \centerline{\footnotesize{(b) Recall}}
\end{minipage}
\begin{minipage}[b]{0.45\linewidth}
  \centering
\includegraphics[width=\linewidth, trim= 40mm 85mm 40mm 85mm]{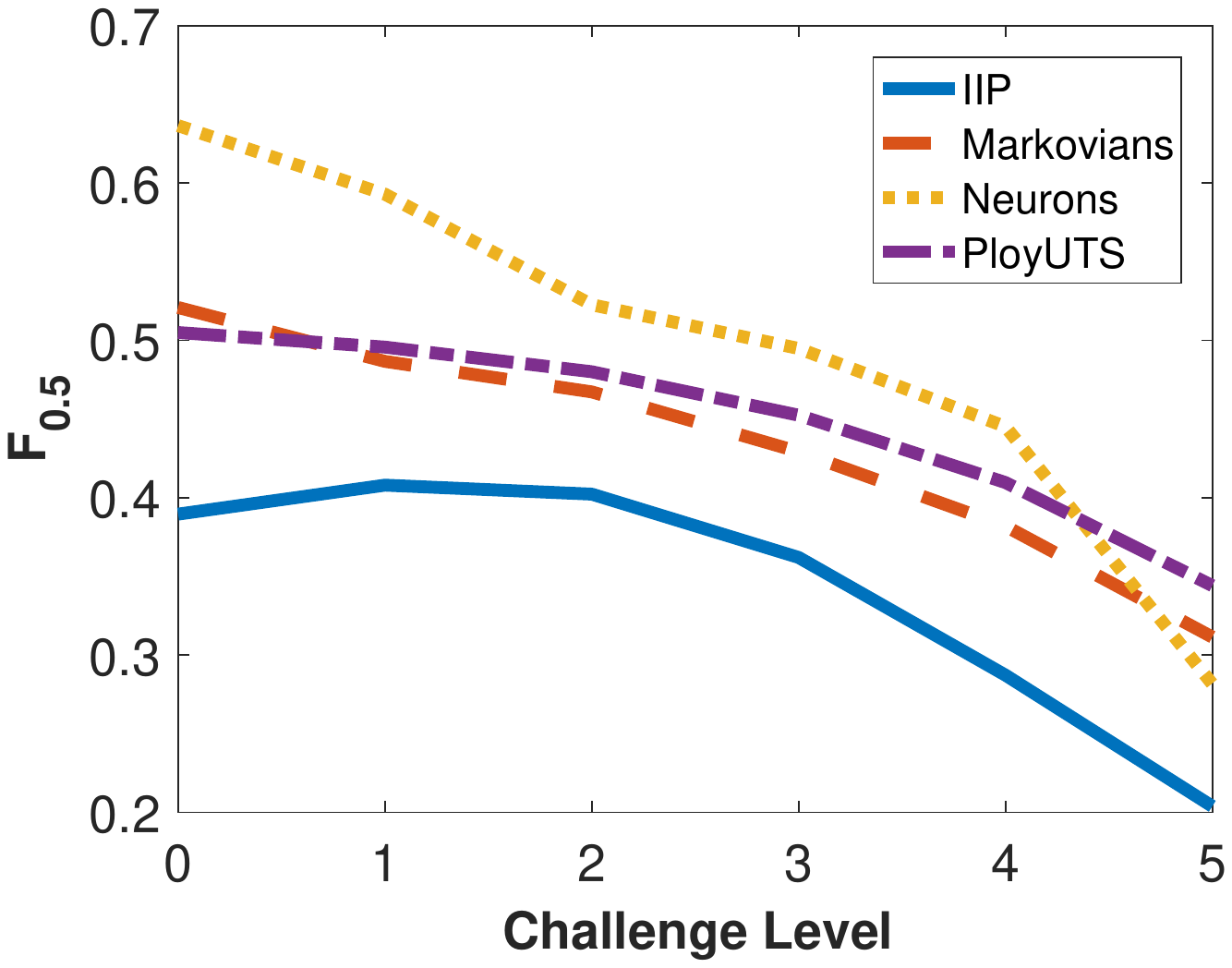}
  \vspace{0.01 cm}
  \centerline{\footnotesize{(c) $F_{0.5}$} }
\end{minipage}
\begin{minipage}[b]{0.45\linewidth}
  \centering
\includegraphics[width=\linewidth, trim= 40mm 85mm 40mm 85mm]{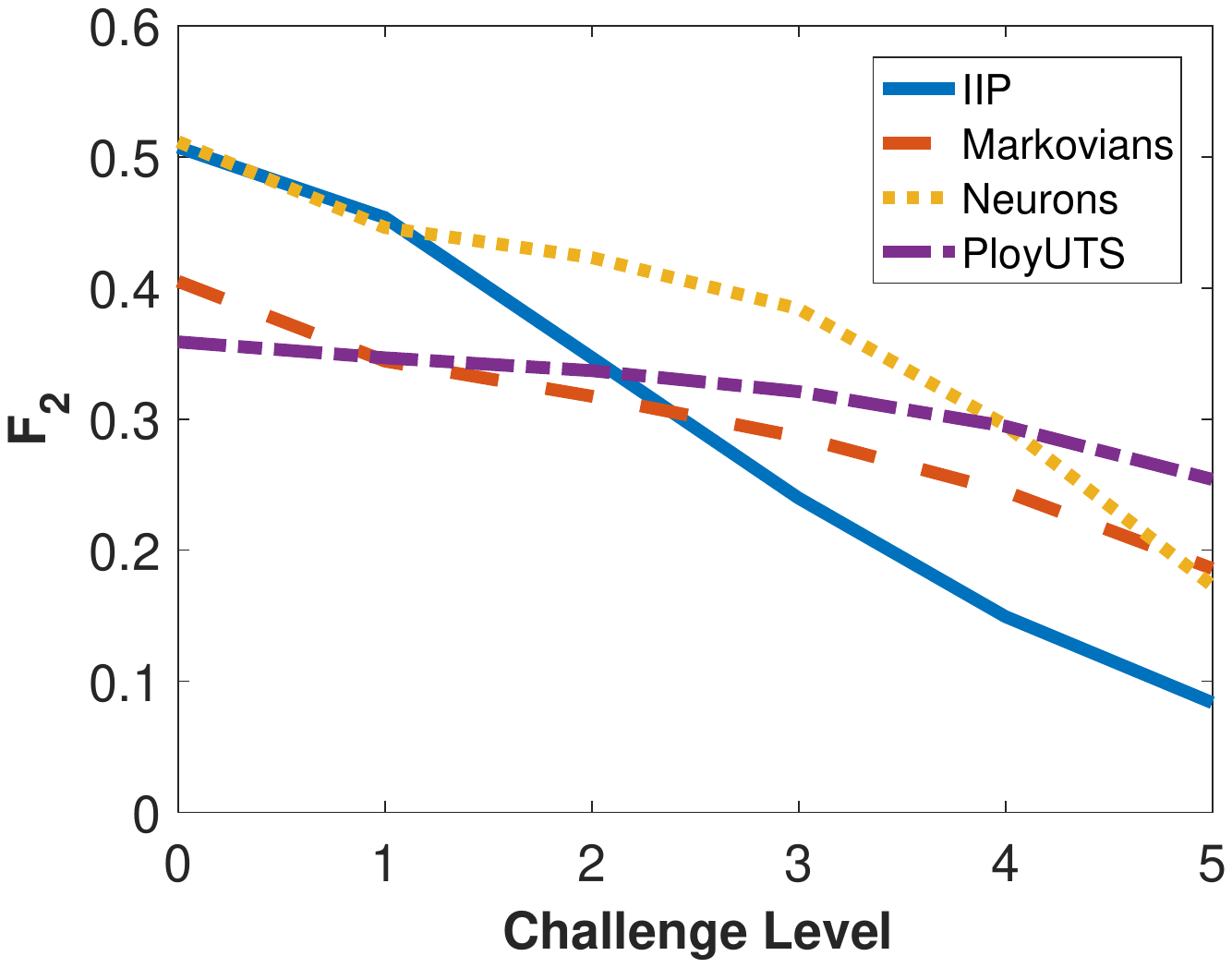}
  \vspace{0.01 cm}
  \centerline{\footnotesize{(d) $F_{2}$   } }
\end{minipage}
\caption{Performance versus challenge levels.}
\label{fig:perfromanceVsLevel}
\end{figure*}

\subsection{Results}
Overall performance of competing teams in each category is summarized in Table. \ref{tab:results}. Higher score corresponds to better performance. The algorithm submitted by team \texttt{Neurons} outperforms all other submissions across all metrics as given in Table \ref{tab:results}. In terms of $Precision$, teams \texttt{PolyUTS} and \texttt{Markovians} follow team \texttt{Neurons}. In all other metrics, team \texttt{PolyUTS} is the second and team \texttt{Markovians} is the third. Team \texttt{IIP} is the fourth best performing algorithm in terms of all the performance metrics in the IEEE VIP Cup 2017. When all performance metrics are considered for ranking, team \texttt{Neurons} is first, team \texttt{PolyUTS} is second, team \texttt{Markovians} is third, and team \texttt{IIP} is fourth.

\begin{figure*}[htbp!]
\centering
\begin{minipage}[b]{0.45\linewidth}
  \centering
\includegraphics[width=\linewidth, trim= 40mm 85mm 40mm 85mm]{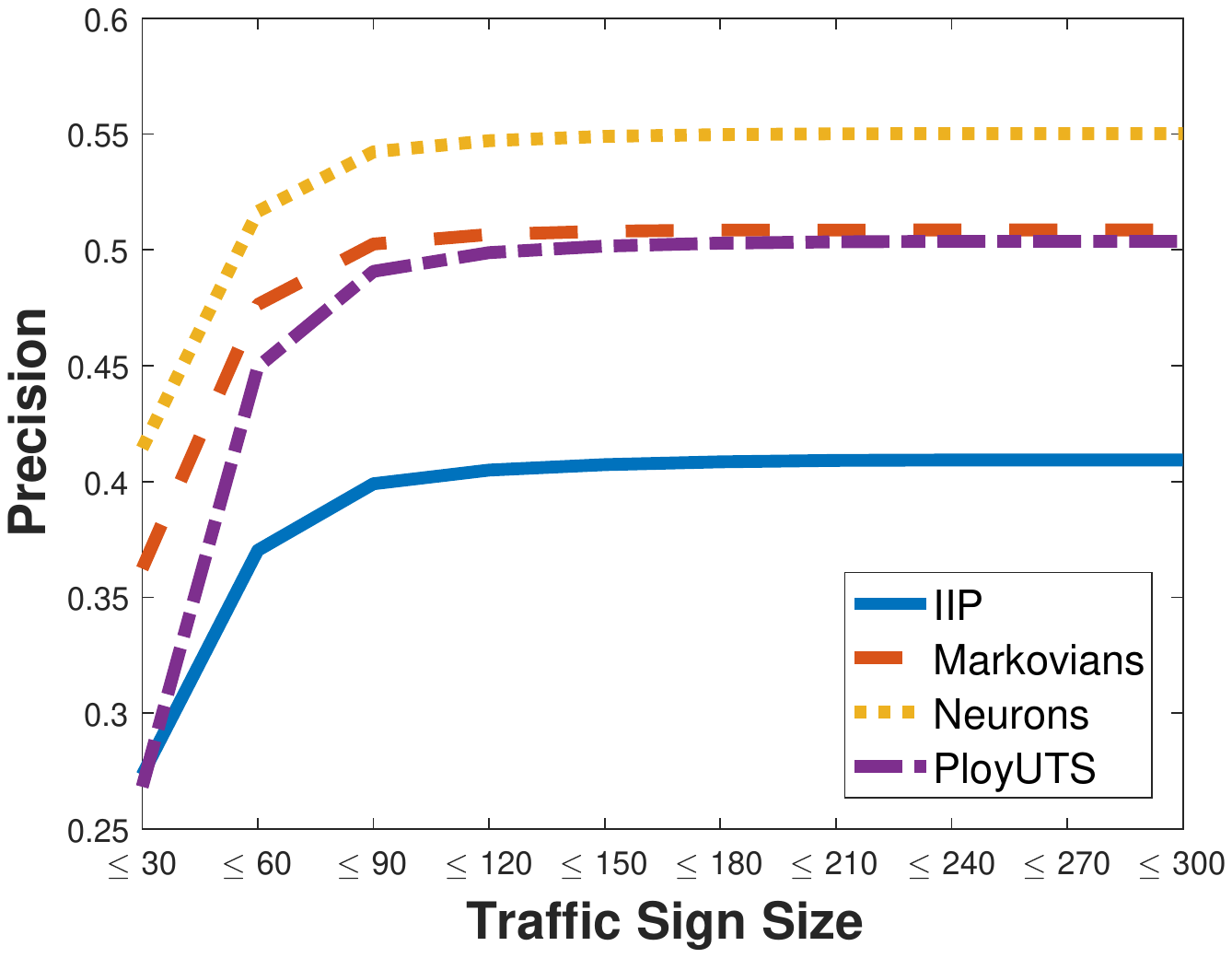}
  \vspace{0.01cm}
  \centerline{\footnotesize{(a)Precision}}
\end{minipage}
\begin{minipage}[b]{0.45\linewidth}
  \centering
\includegraphics[width=\linewidth, trim= 40mm 85mm 40mm 85mm]{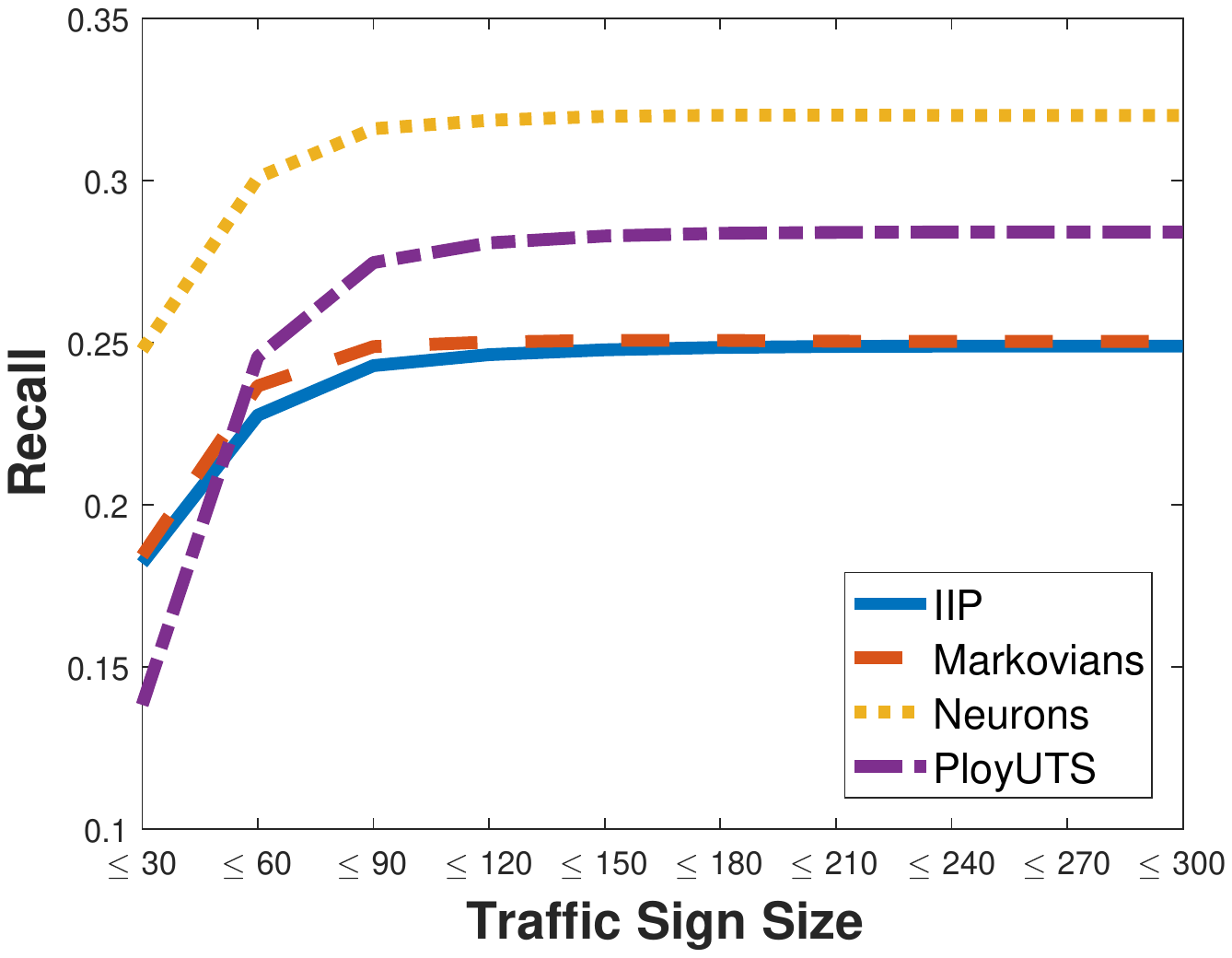}
  \vspace{0.01cm}
  \centerline{\footnotesize{(b) Recall}}
\end{minipage}
\begin{minipage}[b]{0.45\linewidth}
  \centering
\includegraphics[width=\linewidth, trim= 40mm 85mm 40mm 85mm]{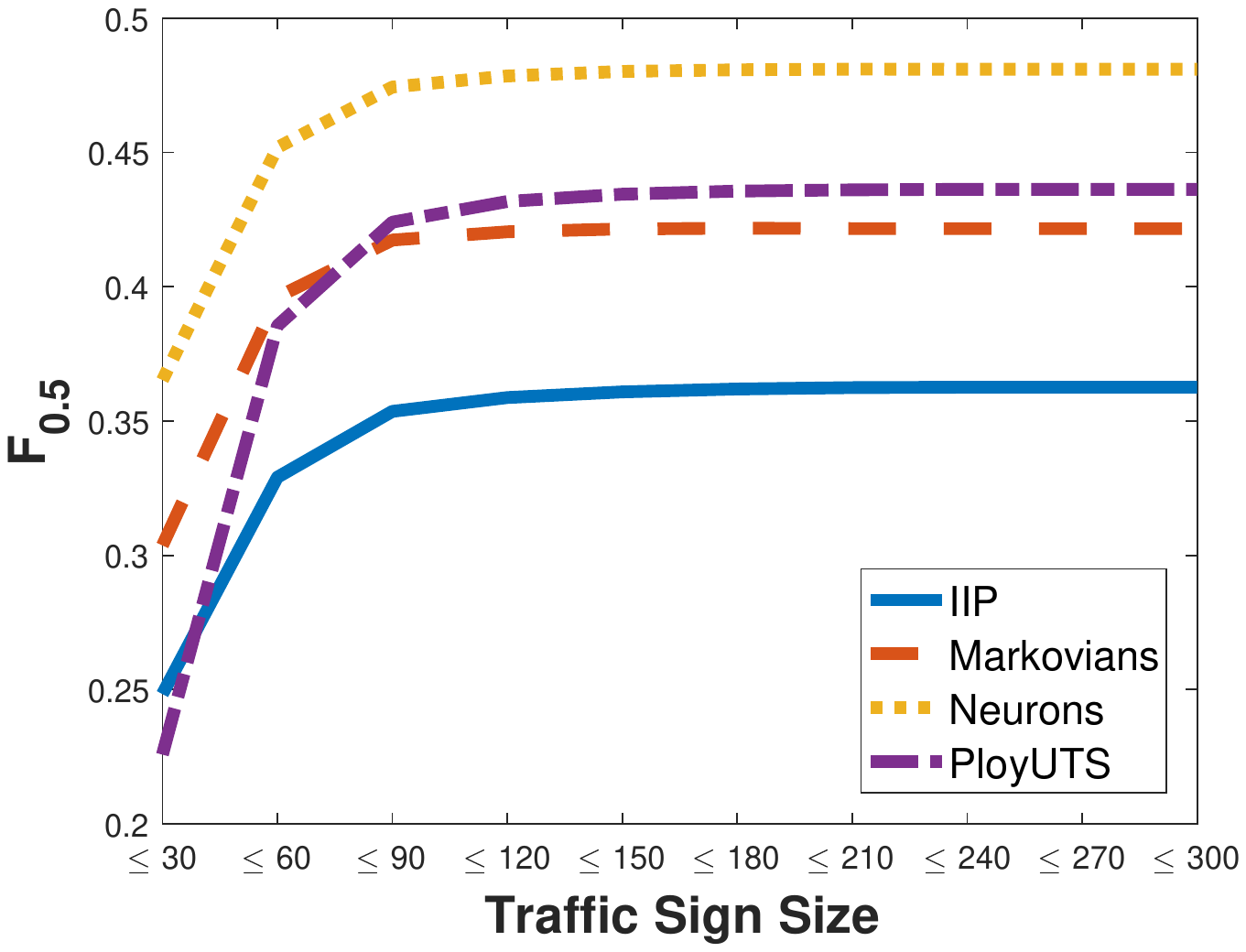}
  \vspace{0.01 cm}
  \centerline{\footnotesize{(c) $F_{0.5}$} }
\end{minipage}
\begin{minipage}[b]{0.45\linewidth}
  \centering
\includegraphics[width=\linewidth, trim= 40mm 85mm 40mm 85mm]{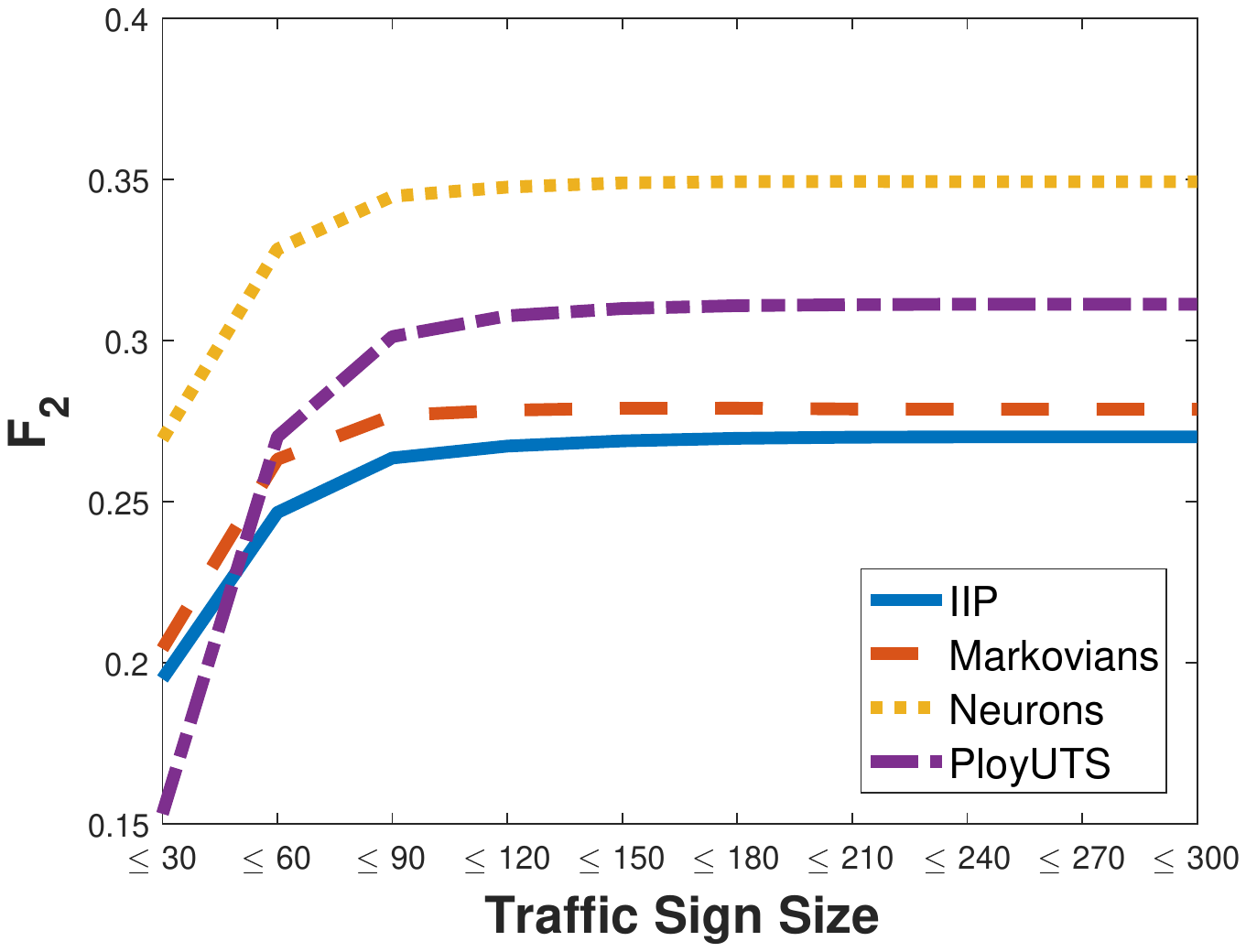}
  \vspace{0.01 cm}
  \centerline{\footnotesize{(d) $F_{2}$   } }
\end{minipage}
\caption{Performance versus sign size.}
\label{fig:perfromanceVsSize}
\end{figure*}

In addition to ranking submitted algorithms, we also analyzed the performance of top four methods with respect to challenge level, traffic sign size, and challenge type. The performance of algorithms versus challenge levels is shown in Fig. \ref{fig:perfromanceVsLevel} in which y-axis corresponds to performance metric values and x-axis corresponds to challenge levels. In terms of $Recall$ and $F_2$ score, algorithmic performance decreases as challenge levels increase. In terms of $Precision$, team \texttt{Neurons} follows a piece-wise linear decrease, team \texttt{PolyUTS} and team \texttt{Markovians} follow a parabolic decrease, and team \texttt{IIP} follows a linear decrease after level $3$. In terms of $F_{0.5}$ score, all metrics follow a monotonically decreasing behavior after level $1$. Team \texttt{Neurons} and team \texttt{IIP} follow a piece-wise linear behavior whereas team \texttt{PolyUTS} and team \texttt{Markovians} follow a parabolic behavior. Aforementioned results indicate that training procedure of team \texttt{IIP} leads to highest $Precision$ for sequences close to level $3$ and highest $F_{0.5}$ for sequences close to level $1$. In all other scenarios, performance of algorithms degrade as challenge level increases.

Varying size of traffic signs can be considered as another challenge source, whose level increases as sign size decreases. We report the performance of algorithms versus traffic sign size in Fig. \ref{fig:perfromanceVsSize} in which y-axis corresponds to performance values and x-axis corresponds to the size limit. Performance of algorithms were calculated for traffic signs whose maximum dimension (width or height) were smaller than the indicated size limit in the x-axis. As x-axis value increases, size of the signs included in the evaluation also increases, which makes is easier to localize and recognize signs. Therefore, we can consider x-axis values as indicator of challenge level, which increases as x-values get closer to the origin. All four algorithms follow a monotonically decreasing behavior with respect to size-based challenge level. Team \texttt{Neurons} leads to highest performance for all size levels in terms of all categories. For all size categories other than size limit $30$, team \texttt{IIP} leads to lowest performance among all four methods. Team \texttt{PolyUTS} leads team \texttt{Markovians} in all the categories other than $Precision$ when size limit is greater than $90$. When size limit is $60$ or less, team \texttt{Markovians} outperforms team \texttt{PolyUTS}.

\begin{figure*}[htbp!]
\centering
    \begin{subfigure}[b]{0.4\linewidth}
        \includegraphics[width=1\linewidth, trim= 40mm 85mm 40mm 85mm]{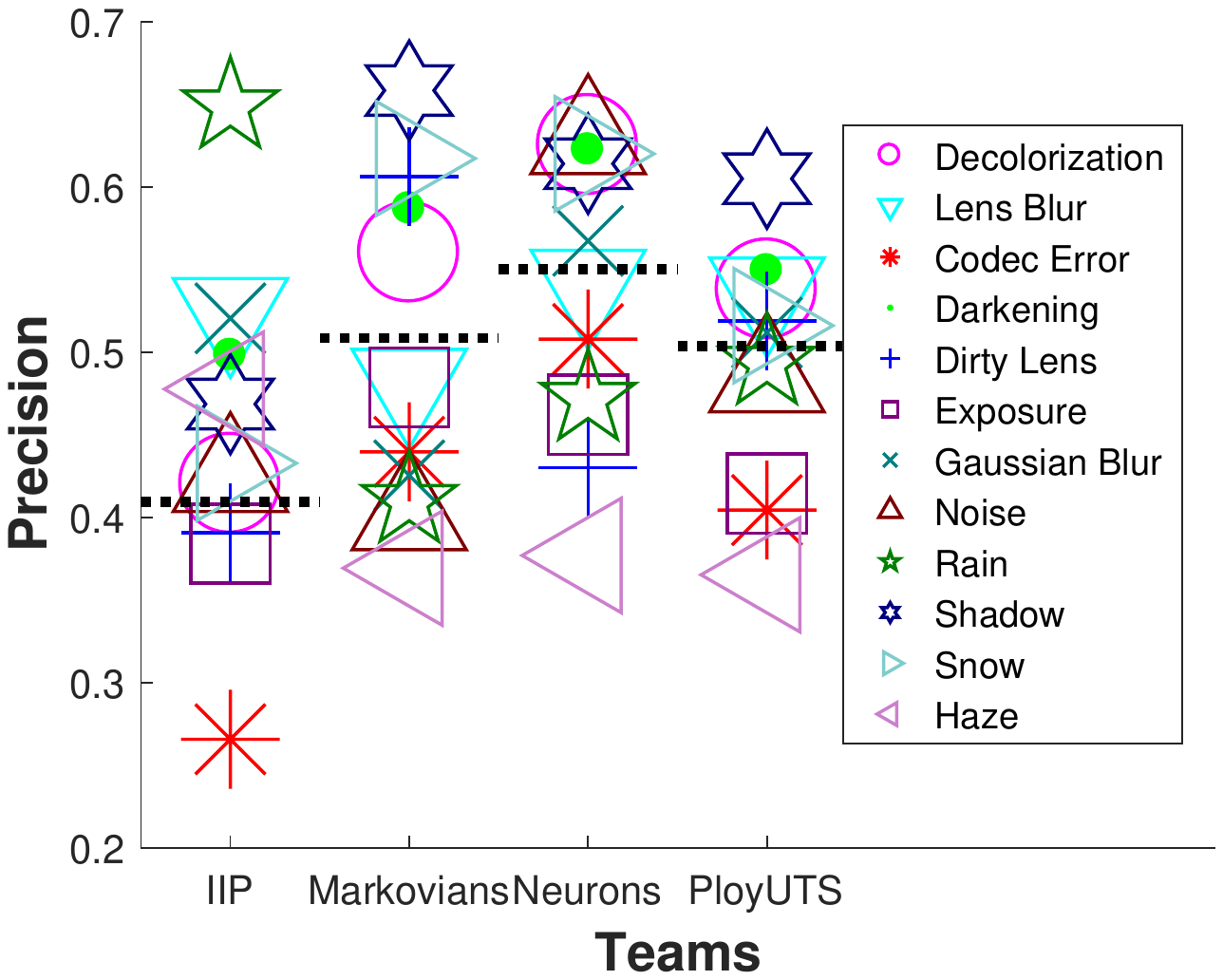}
        \caption{Precision}
    \end{subfigure}
    \begin{subfigure}[b]{0.4\linewidth}
        \includegraphics[width=1\linewidth, trim= 40mm 85mm 40mm 85mm]{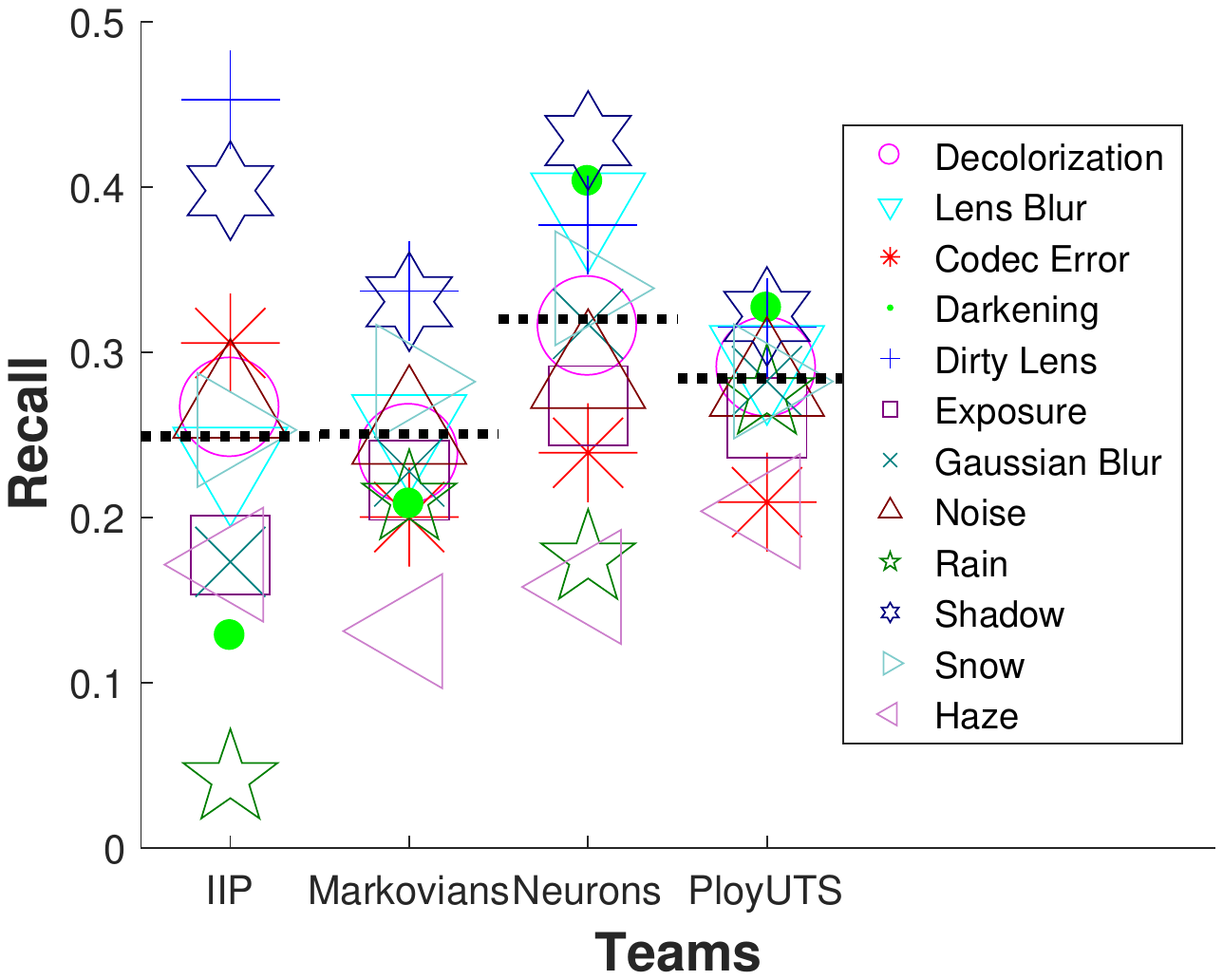}
        \caption{Recall}
    \end{subfigure}
    \begin{subfigure}[b]{0.4\linewidth}
        \includegraphics[width=1\linewidth, trim= 40mm 85mm 40mm 88mm]{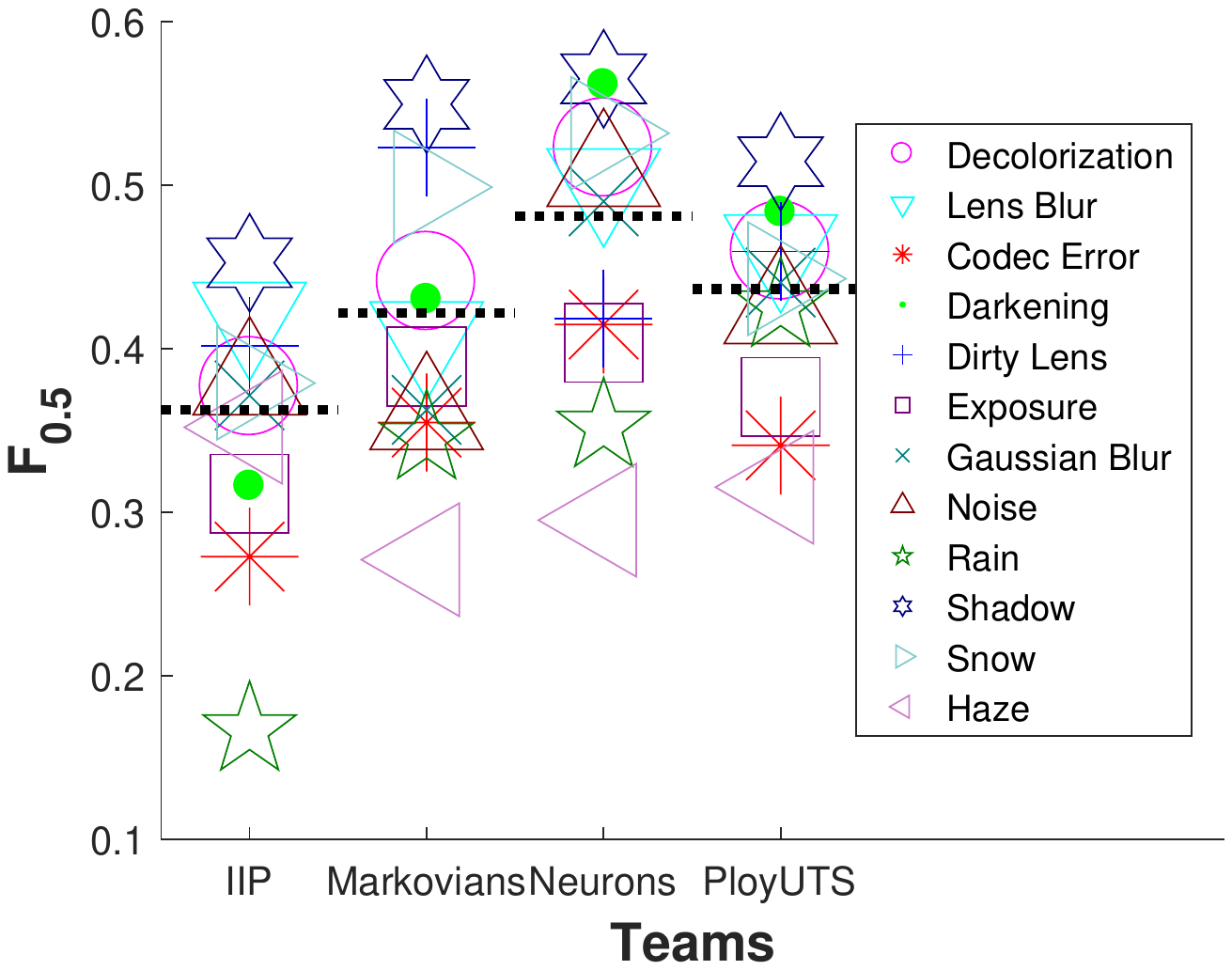}
        \caption{$F_{0.5}$}
    \end{subfigure}
    \begin{subfigure}[b]{0.4\linewidth}
        \includegraphics[width=1\linewidth, trim= 40mm 85mm 40mm 88mm]{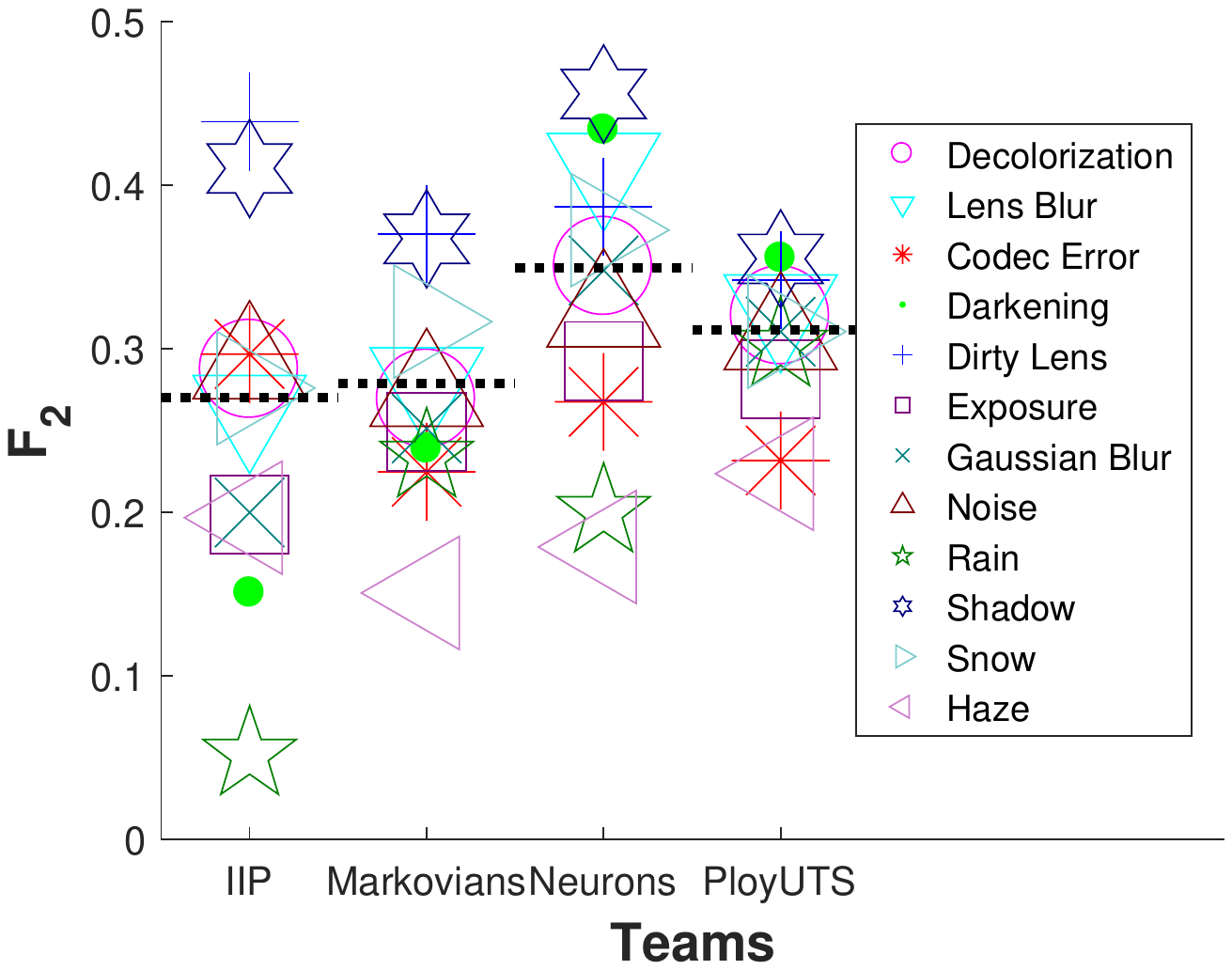}
        \caption{$F_{2}$}
    \end{subfigure}
\caption{Performance versus challenge types.}
\label{fig:perfromanceVsTypes}
\vspace{-4 mm}
\end{figure*}

Previously, we have investigated the effect of challenge level on the algorithmic performance. To understand the effect of challenging conditions, we also need to analyze the effect of challenge types. We report the performance of algorithms versus challenge type in Fig. \ref{fig:perfromanceVsTypes} in which the y-axis corresponds to performance values and the x-axis corresponds to the top four algorithms from participating teams. Performance values corresponding to different challenge categories are represented with different colors and shapes. Overall performance of each team is shown with a horizontal dotted line. Performance change with respect to challenge types is more significant in team \texttt{IIP} compared to other algorithms. Specifically, overall performance of team \texttt{IIP} significantly degrades because of codec error in $Precision$ and because of rain in all other metrics, which leads to inferior overall performance compared to top three performing methods. Team \texttt{Markovians} has a high performance variation in $Precision$ and $F_{0.5}$ whereas their performance variation is low in $Recall$ and $F_2$. When low performance variation in $Recall$ and $F_2$ is combined with the lowest performance in haze category,  team \texttt{Markovians} performs similar to team \texttt{IIP} and both teams underperform compared to team \texttt{PolyUTS} and teams \texttt{Neurons}. Even though team \texttt{Neurons} has high performance variation in each metric, their relatively superior performance in the majority of challenge types makes them the top performing algorithm.

\begin{figure}[htbp!]
\vspace{-3 mm}
\centering
\includegraphics[width=\linewidth]{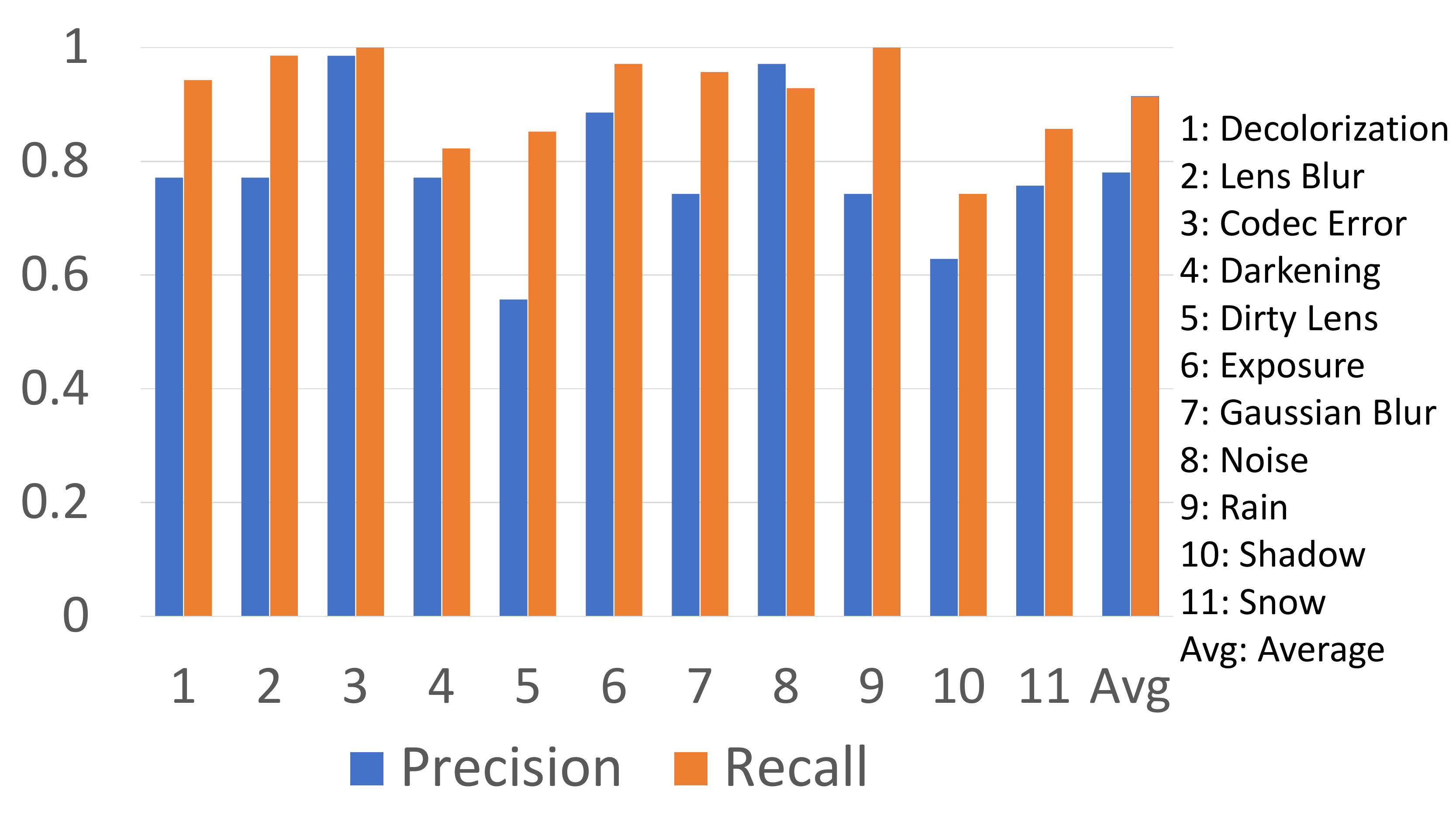}
\caption{Average Spearman correlation between detection performances in real-world and synthesized environments.}
\vspace{-5mm}
\label{fig:real_unreal_corr}
\end{figure}

Aforementioned experiments focus on the detection performance in the test set, which includes a combination of real-world and synthesized data. However, to understand the effect of challenging conditions in different environments, we can split the test set into real and synthesized sets and compare their performances. To measure the correlation between detection performances in real-world data and synthesized data, we computed the average Spearman correlation for $Precision$ and $Recall$ per challenging conditions as provided in Fig.~\ref{fig:real_unreal_corr}. On average, Spearman correlation is 0.915 for $Recall$ and 0.781 for $Precision$ when real and synthesized results were compared. A high correlation between real-world and synthesized environment performances indicate that testing robustness of algorithms under challenging conditions in a simulator environment can be used to predict the robustness of algorithms in similar real-world scenarios. 

\section{Conclusion}
We performed a comprehensive analysis of publicly available traffic sign datasets. Based on our analysis, we observed that existing datasets are limited in terms of challenging environmental conditions and they lack metadata of challenge conditions and levels. Moreover, existing datasets are also limited in terms of size and annotated frames that are continuous. Finally, it is not possible to determine the relationship between individual environmental conditions and algorithmic performance because multiple conditions change simultaneously. To overcome the shortcomings in the traffic sign literature, we introduced the \texttt{CURE-TSD} dataset and organized the IEEE VIP Cup 2017 to benchmark algorithms under challenging conditions. All top performing algorithms are based on deep architectures whose performance degrade significantly under challenging conditions. Therefore, conducted study articulates the urgent need for developing more robust algorithms that can adapt to the variations in environmental conditions. To understand the effect of challenging conditions in different environments, we compared the detection performance variation in real-wold data and synthesized data. Experimental results showed that there is a high correlation between real-world and synthesized performance variations, which indicates that testing robustness with simulator data can be used to predict the real-world reliability under similar scenarios.


\ifCLASSOPTIONcaptionsoff
  \newpage
\fi



\begin{IEEEbiography}
    [{\includegraphics[width=1.0in]{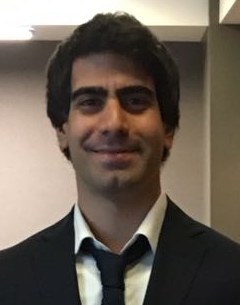}}]{Dogancan Temel} received his
M.S. and Ph.D. degrees in electrical and computer engineering (ECE) from the Georgia
Institute of Technology (Georgia Tech) in 2013 and 2016, respectively, where he is currently a postdoctoral fellow. During his PhD, he received Texas Instruments Leadership Universities Fellowship for four consecutive years. He was the recipient of the Best Doctoral Dissertation Award from the Sigma Xi honor society, the Graduate Research Assistant Excellence Award from the School of ECE, and the Outstanding Research Award from the Center for Signal and Information Processing at Georgia Tech. His research interests are focused on human and machine vision.

\end{IEEEbiography}


\begin{IEEEbiography}
        [{\includegraphics[width=1.0in]{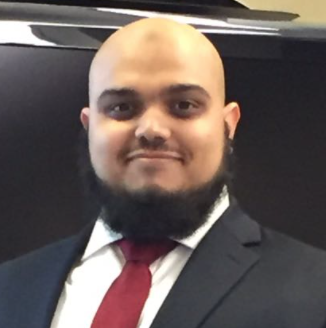}}]{Tariq Alshawi} received an M.S. degree with a minor in mathematics in 2013 from University of Michigan, Ann Arbor and a PhD  degree in 2018 from the school of Electrical and Computer Engineering in Georgia Institute of Technology, Atlanta. Currently, he is a professor in the Electrical Engineering Department of King Saud University. His research interests are image and video processing and human vision system.
\end{IEEEbiography}


\begin{IEEEbiography}
    [{\includegraphics[width=1in]{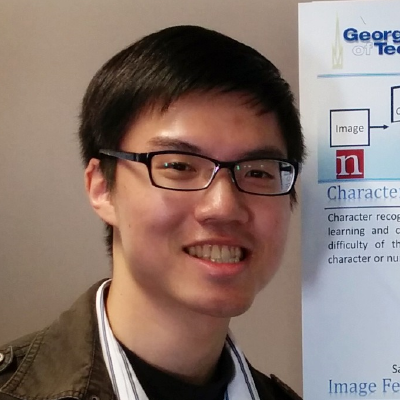}}]{Min-Hung Chen}  received an M.S. degree in 2012 from National Taiwan University. He worked as a research assistant in Academia Sinica, Taiwan in 2013. He is currently working towards PhD in the school of Electrical and Computer Engineering in Georgia Institute of Technology, Atlanta. During his PhD, he received Ministry of Education Technologies Incubation Scholarship from Taiwan for three years. His research interests include computer vision, deep learning, image and video processing. Recently, Min-Hung Chen is investigating the temporal dynamics through deep learning techniques.
\end{IEEEbiography}


\begin{IEEEbiography}
    [{\includegraphics[width=1in]{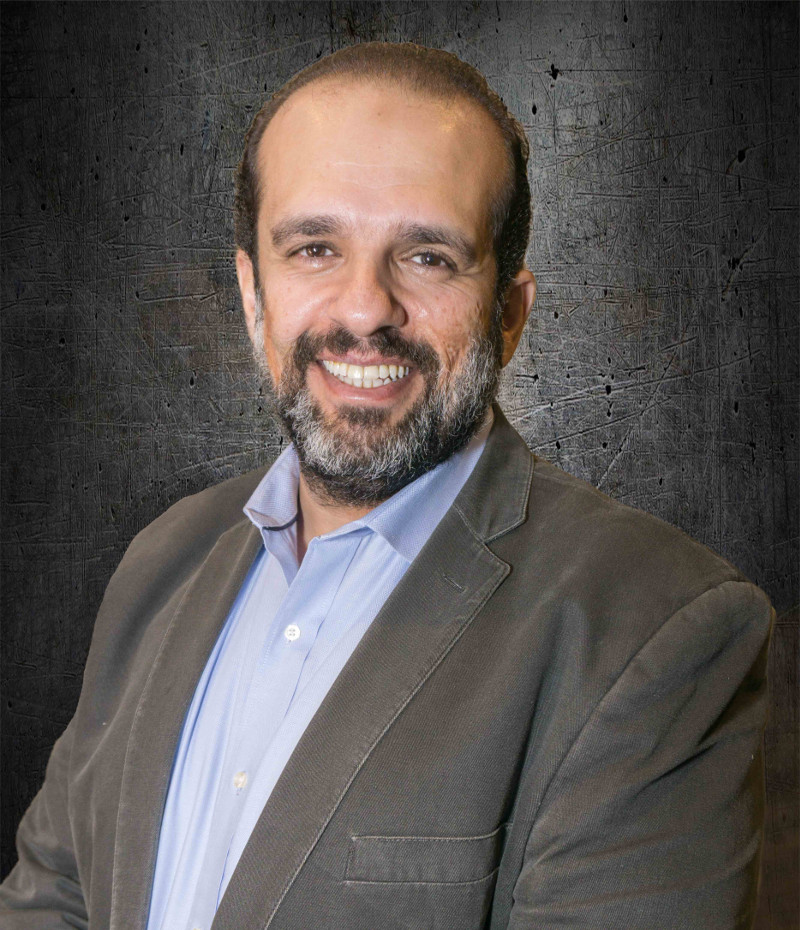}}]{Ghassan AlRegib} is currently a Professor with the School of Electrical and Computer Engineering, Georgia Institute of Technology.  He was a recipient of the ECE Outstanding Graduate Teaching Award in 2001 and both the CSIP Research and the CSIP Service Awards in 2003, the ECE Outstanding Junior Faculty Member Award, in 2008, and the 2017 Denning Faculty Award for Global Engagement. His research group, the Omni Lab for Intelligent Visual Engineering  and Science (OLIVES) works on research projects related to machine learning, image and  video processing, image and video understanding, seismic interpretation, healthcare intelligence, machine learning for ophthalmology, and video analytics. He participated in several service activities within the IEEE including the organization of the First IEEE VIP Cup (2017), Area Editor for the IEEE Signal Processing Magazine, and the Technical Program Chair of GlobalSIP’14 and ICIP’20. He has provided services and consultation to several firms, companies, and international educational and Research and Development organizations. He has been a witness expert in a number of patents infringement cases. 
\end{IEEEbiography}

\newpage

\begin{figure*}[htbp!]
\centering
\begin{minipage}[b]{\linewidth}
  \centering
\includegraphics[width=0.9\linewidth]{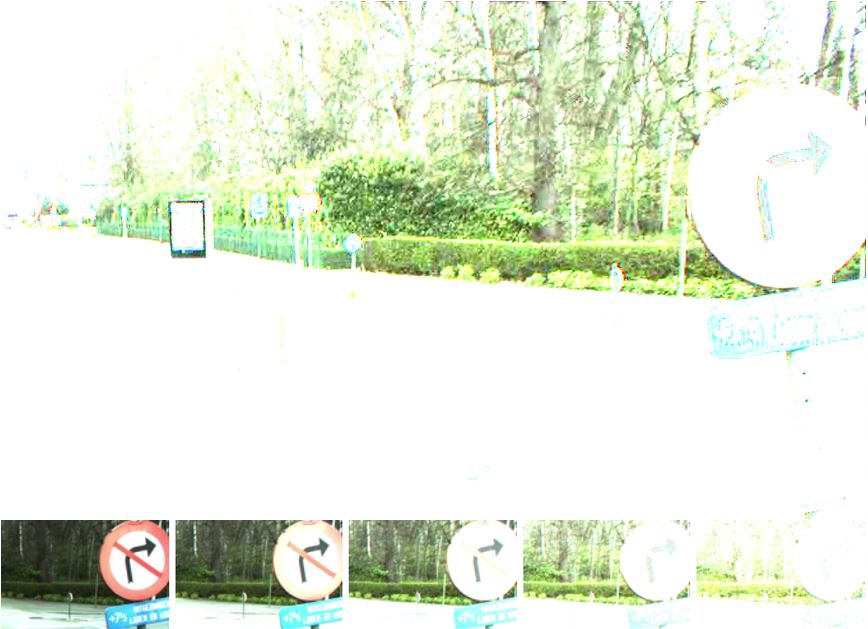}
  \vspace{0.1cm}
  \centerline{\footnotesize{(a) Exposure condition in the CURE-TSD dataset. }}
  \vspace{0.1cm}
\end{minipage}
\begin{minipage}[b]{\linewidth}
  \centering
\includegraphics[width=0.9\linewidth]{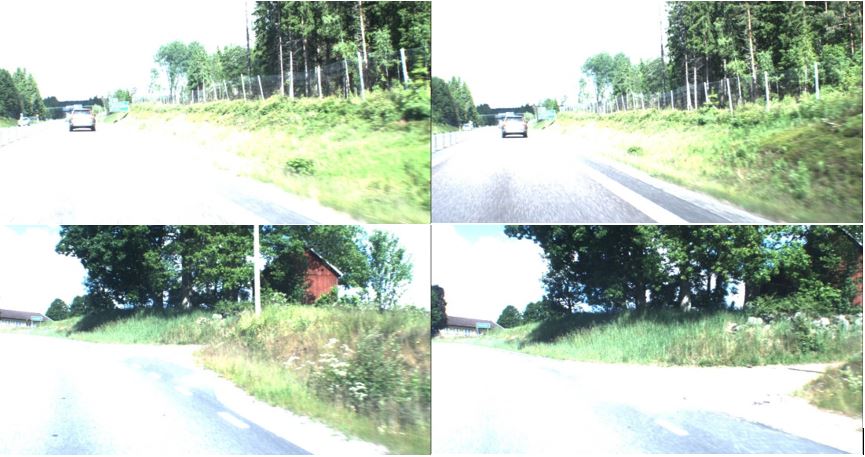}
  \vspace{0.1cm}
  \centerline{\footnotesize{(b) Exposure condition in the STS  dataset \cite{Larsson2011}. }}
  \vspace{0.1cm}
\end{minipage}
\caption{Exposure condition in the CURE-TSD dataset and the STS datset. Different challenge levels corresponding to the same scene are shown for the CURE-TSD dataset whereas different scene examples are shown for the STS dataset.}
\label{fig:challenges_exposure}
\end{figure*}

\begin{figure*}[htbp!]
\centering
\begin{minipage}[b]{\linewidth}
  \centering
\includegraphics[width=0.9\linewidth]{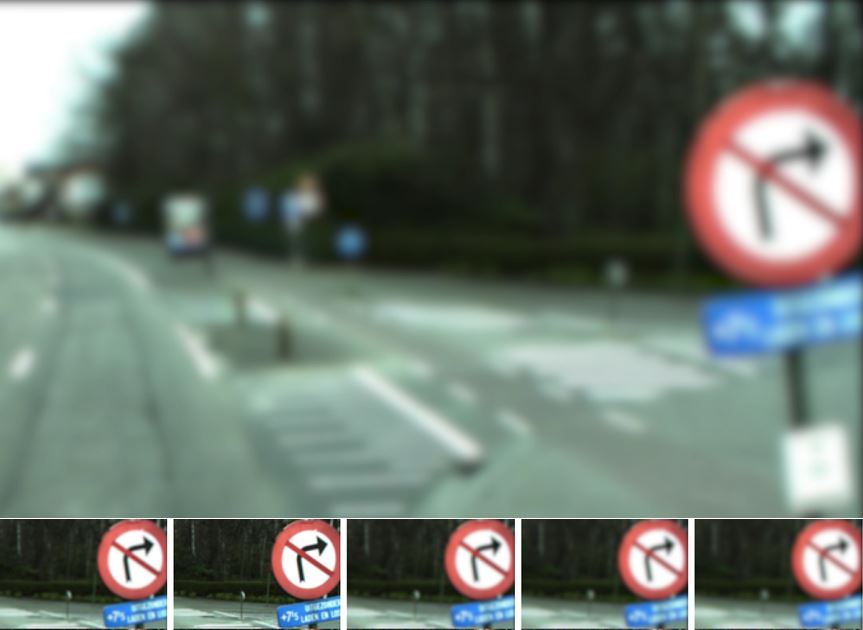}
  \vspace{0.1cm}
  \centerline{\footnotesize{(a) Blur condition in the CURE-TSD dataset. }}
  \vspace{0.1cm}
\end{minipage}
\begin{minipage}[b]{\linewidth}
  \centering
\includegraphics[width=0.9\linewidth]{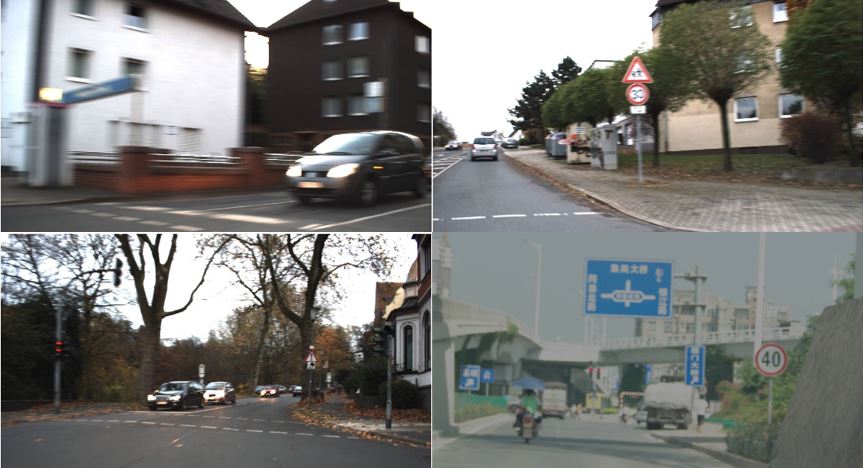}
  \vspace{0.1cm}
  \centerline{\footnotesize{(b) Blur condition in the GTSDB  dataset \cite{Houben2013} and the CCTSDB \cite{Zhang2017} dataset. }}
  \vspace{0.1cm}
\end{minipage}
\caption{Blur condition in the CURE-TSD dataset, the GTSDB dataset, and the CCTSDB dataset. Different challenge levels corresponding to the same scene are shown for the CURE-TSD dataset whereas different scene examples are shown for the GTSDB dataset and the CCTSDB dataset.}
\label{fig:challenges_exposure}
\end{figure*}

\end{document}